\documentclass[lettersize,journal]{IEEEtran}
\usepackage{amsmath,amsfonts}
\usepackage{algorithmic}
\usepackage{algorithm}
\usepackage{array}
\usepackage[caption=false,font=normalsize,labelfont=sf,textfont=sf]{subfig}
\usepackage{textcomp}
\usepackage{stfloats}
\usepackage{url}
\usepackage{verbatim}
\usepackage{graphicx}
\usepackage{cite}
\usepackage{booktabs}
\usepackage{siunitx}
\usepackage{float}
\begin{document}

\title{POPL-KF: A Pose-Only Geometric Representation-Based Kalman Filter for Point-Line-Based Visual-Inertial Odometry}

\author{Aiping Wang$^{\dagger}$, Zhaolong Yang$^{\dagger}$, Shuwen Chen$^{\dagger}$, and Hai Zhang$^{\ast}$
\thanks{$^{\dagger}$Aiping Wang, Zhaolong Yang, and Shuwen Chen contribute equally to this work.}
\thanks{$^{\ast}$Hai Zhang is the corresponding author.}
\thanks{This work is supported by the National Natural Science Foundation of China (No.~62373031) and the Guizhou Science and Technology Program Project (No.~2023-341) }
\thanks{Aiping Wang, Zhaolong Yang, Shuwen Chen and Hai Zhang are with the School of Automation Science and Electrical Engineering, Beihang University, Beijing 100191, China.(Email:wangaiping@buaa.edu.cn; yangzhaolong@buaa.edu.cn;  chenshuwen@buaa.edu.cn;  zhanghai@buaa.edu.cn.) }}

\markboth{Submitted to IEEE}%
{Shell \MakeLowercase{\textit{et al.}}: A Sample Article Using IEEEtran.cls for IEEE Journals}


\maketitle
\IEEEpubid{
	\begin{minipage}{\textwidth}
		\centering
		\vspace{39pt} 
		\fbox{
			\begin{minipage}{0.9\textwidth} 
				\centering
				\scriptsize 
				\copyright 2026 IEEE. Personal use of this material is permitted. Permission from IEEE must be obtained for all other uses, in any current or future media, including reprinting/republishing this material for advertising or promotional purposes, creating new collective works, for resale or redistribution to servers or lists, or reuse of any copyrighted component of this work in other works.
			\end{minipage}
		}
	\end{minipage}
}

\begin{abstract}
Mainstream Visual-inertial odometry 	
(VIO) systems rely on point features for motion estimation and localization. However, their performance degrades in challenging scenarios. Moreover, the localization accuracy of multi-state constraint Kalman filter (MSCKF)-based VIO systems suffers from linearization errors associated with feature 3D coordinates and delayed measurement updates. To improve the performance of VIO in challenging scenes, we first propose a pose-only geometric representation for line features. Building on this, we develop POPL-KF, a Kalman filter-based VIO system that employs a pose-only geometric representation for both point and line features. POPL-KF mitigates linearization errors by explicitly eliminating both point and line feature coordinates from the measurement equations, while enabling immediate update of visual measurements. We also design a unified base-frames selection algorithm for both point and line features to ensure optimal constraints on camera poses within the pose-only measurement model. To further improve line feature quality, a line feature filter based on image grid segmentation and bidirectional optical flow consistency is proposed. Our system is evaluated on public datasets and real-world experiments,  demonstrating that POPL-KF outperforms the state-of-the-art (SOTA) filter-based methods (OpenVINS, PO-KF) and optimization-based methods (PL-VINS, EPLF-VINS), while maintaining real-time performance.
\end{abstract}

\begin{IEEEkeywords}
Visual-inertial odometry (VIO), pose-only representation, localization, multi-state constraint Kalman filter, line features.
\end{IEEEkeywords}

\section{Introduction}
\IEEEPARstart{V}{isual}-inertial odometry (VIO) is fundamental to the localization and mapping of autonomous driving, virtual reality, and robotic navigation \cite{1zhang2022novel,1adqin2020lins,lim2022uv}. The current mainstream VIO systems can be broadly categorized into optimization-based\cite{8ge2024pipo} and filter-based systems\cite{3geneva2020openvins}. Optimization-based VIO systems can usually achieve higher accuracy through iterative linearization. However, their high computational cost may limit real-time applicability. In contrast, filter-based systems balance accuracy and computational load. They are well-suited for platforms with limited resources.

The Multi-State Constraint Kalman Filter (MSCKF) stands as a quintessential filter-based VIO systems \cite{6mourikis2007multi}. In its measurement update pipeline, feature coordinates are triangulated once the features are lost or reach their maximum tracking length, and are subsequently projected onto the image planes to construct the measurement equations. Since feature coordinates are not maintained in the system state, MSCKF employs null-space projection to eliminate them from the measurement equations. However, the linearization errors persist in the innovation and Jacobian matrices, which consequently degrade state estimation accuracy. Moreover, delayed updates may lead to accumulated errors, which pose challenges to maintaining localization accuracy for consumer-grade IMUs. Equally important, if feature triangulation fails, the measurement equation cannot be constructed, rendering numerous short-tracked features ineffective in the estimation process.

Cai et al. pioneered a pose-only representation for point features and prove its theoretical equivalence to two-view geometry\cite{7cai2021pose}. This approach decouples camera poses from feature positions and provides an analytical solution for 3D coordinates reconstruction. Recent works \cite{8ge2024pipo, 9wang2025po} incorporate the pose-only representation into optimization-based and filter-based VIO systems, respectively. In this way, the system explicitly eliminates point features from the measurement model, thereby mitigating linearization errors. Moreover, it eliminates the dependence on feature triangulation, and the measurement equation no longer depends on the success of triangulation. As a result, immediate updates can be performed once sufficient camera poses are available for the current point feature. These studies demonstrate that the pose-only representation significantly enhances localization accuracy. 

Currently, point-feature-based VIO systems have reached an advanced stage and can achieve high-precision localization in standard environments. However, in challenging conditions such as low-texture or motion-blurred environments, their performance significantly decreases or even diverges due to the inherent instability of point features. Notably, such environments are often man-made and typically contain abundant line features with rich structural information\cite{7ad3lin2024flm}. Compared with point features, line features contain richer structural and geometric information and exhibit greater robustness to challenging factors. In recent years, incorporating line features into VIO systems to improve robustness and localization accuracy has received increasing attention\cite{10wei2021point}.While most existing frameworks rely on the LSD detector and LBD descriptor, this combination is computationally prohibitive for real-time applications \cite{7adavon2008lsd,9he2018pl,7Ad2zhang2013efficient}. EDlines is a real-time line feature extraction algorithm\cite{5xu2022eplf}. In this work, we propose an enhanced version of EDLines to increase the reliability and quality of extracted line features.

Regarding point feature representation, three-dimensional (3D) Cartesian coordinates relative to the world or an anchored frame constitute the most general form. To achieve higher computational efficiency and numerical robustness, the 1D inverse depth parameterization was proposed and widely used in VIO systems\cite{forster2014svo,montiel2006unified}. The Plücker representation is commonly adopted in VIO for line feature modeling, but it suffers from over-parameterization\cite{9he2018pl,10fu2009pl}. To solve this issue, the four degrees-of-freedom (4-DoF) closest point parameterization\cite{yang2019visual,11yang2019aided} and the orthogonal parameterization\cite{zou2019structvio} have been proposed for constructing the measurement equations. Despite continuous improvements in line feature extraction, matching, and parameterization methods, linearization errors and delayed updates remain inherent challenges, ultimately constraining the system’s localization accuracy. 

To address these limitations, a preliminary exploration of the pose-only representation for line features was presented in our previous work\cite{zhanglong}. This approach mitigates linearization errors and delayed updates inherent in the traditional MSCKF. In this work, we substantially extend that foundation to develop POPL-KF, a Kalman filter-based VIO system that employs a unified pose-only geometric representation for both point and line features. The main contributions of this paper are summarized as follows: 

1)	A rigorous pose-only geometric representation for line features is proposed, which is derived from the classical multi-view geometry of line features and is physically interpretable.

2)	We propose a novel pose-only representation for a point-line-based VIO system. To the best of our knowledge, this is the first time where both point and line features adopt pose-only measurement equations. These equations are directly constructed based on the pose of selected base-frames, which explicitly eliminate the feature coordinates from the measurement equations and enable immediate state updates using visual measurements. Furthermore, a unified base-frames selection strategy suitable for both point and line features is designed. 

3)	We develop a line feature filtering method that integrates image gridding with bidirectional optical flow consistency to improve line feature tracking quality and enhance the accuracy of the proposed VIO system. 

4)	To validate the performance of the proposed system, we perform comprehensive evaluations on the EuRoC and KAIST public datasets, as well as on real-world experiments. Experimental results demonstrate that our system outperforms other state-of-the-art (SOTA) VIO systems in localization accuracy and robustness.

The rest of this article is structured as follows. In Section \ref{rel}, we present related work on VIO. Section \ref{pre} introduces the preliminaries of filter-based VIO, Section \ref{ove} gives an overview of the system.and we present the details of the POPL-KF in Section \ref{poseonly}. Experimental results are provided in Section \ref{exp}. Finally, we conclude the paper in Section \ref{con}.
\section{RELATED WORK}
\label{rel}
VIO is a widely studied topic that aims to estimate precise 6-DoF poses by fusing visual and inertial measurements. Based on the fusion strategy, VIO methods can be broadly categorized into filter-based and optimization-based systems.
\subsection{Optimization-Based VIO}
Optimization-based VIO estimate camera poses by solving a nonlinear least-squares problem through bundle adjustment (BA) or graph optimization over a set of observations\cite{gomez2019pl}. A representative system is OKVIS \cite{12leutenegger2015keyframe}, which employs a keyframe-based sliding window to estimate camera poses. VINS-Mono is widely recognized as a benchmark for its competitive localization accuracy. However, its exclusive reliance on point features may lead to degraded pose estimation in challenging environments \cite{4qin2018vins}. PL-VIO\cite{9he2018pl} employs the LSD algorithm and the LBD descriptor for line extraction and matching, but fails to maintain real-time performance. PL-VINS addresses this issue by integrating an enhanced LSD-based line feature extraction algorithm, enabling real-time operation\cite{10fu2009pl}. The EDLines detector \cite{akinlar2011edlines} achieves comparable line detection performance to LSD while offering significantly higher computational efficiency. Building upon PL-VINS, Xu et al. propose EPLF-VINS\cite{5xu2022eplf}, which incorporates an improved EDLines-based line detection module. EPLF-VINS is one of SOTA visual inertial integrated navigation systems. 

The traditional BA requires joint nonlinear optimization of camera poses and feature coordinates, which introduces estimation errors and increases computational cost \cite{campos2021orb,xu2023plpl}. To address this issue, Cai et al. proposed a pose-only three-view geometry theory for point features and subsequently developed pose-only bundle adjustment (PA)\cite{7cai2021pose}. This method optimizes only the camera poses, effectively eliminating errors in feature coordinate estimation. Ge et al. integrated PA into ORB-SLAM3 and proposed PIPO-SLAM, which achieves improved computational and storage efficiency\cite{8ge2024pipo}. However, since BA performs iterative optimization of feature coordinates, resulting in low estimation errors, PIPO-SLAM achieves only marginal improvements in positioning accuracy compared to ORB-SLAM3.

Benchmark tests show that optimization-based VIO systems provide higher accuracy but also incur higher computational costs. The introduction of line features further increases the computational load, making it difficult to maintain real-time performance on resource-limited platforms. In contrast, filter-based VIO imposes lower computational demands and generally demonstrates more consistent performance across diverse datasets.

\subsection{Filter-based VIO}
Unlike optimization-based methods, filter-based VIO linearizes the system state separately during propagation and update, which makes it computationally more efficient. Early filter-based VIO frameworks typically employed the Extended Kalman Filter (EKF)\cite{li2013high}, in which feature coordinates were augmented into the system state. However, as the number of features increases, the state dimension grows significantly, leading to higher computational complexity. Moreover, such approaches suffer from substantial linearization errors.

MSCKF is one of the most representative filter-based VIO systems.  It only augments the camera pose to the state, effectively mitigating the state dimensionality explosion and achieving comparable localization accuracy to other methods. Huang et al. proposed the First Estimate Jacobian (FEJ) method and integrated it into MSCKF to prevent observability degradation caused by state inconsistency\cite{huang2009first}. Wu et al. introduced a square root inverse version of MSCKF to improve computational efficiency and numerical stability, enabling it to run on resource-limited devices without sacrificing estimation accuracy\cite{wu2015square}. The Vijay Kumar Lab released an open-source stereo version of MSCKF, named S-MSCKF\cite{sun2018robust}, which significantly improves robustness while maintaining computational efficiency comparable to SOTA monocular methods. ROVIO\cite{bloesch2017iterated} introduces a robocentric formulation and iterated filtering to refine the accuracy of linearization points. Qiu et al. proposed the Lightweight-Hybrid MSCKF, which models Zero Velocity Updates (ZUPT) as a concise closed-form measurement update to achieve a favorable trade-off between computational efficiency and localization accuracy\cite{xiaochen2020lightweight}. OpenVINS\cite{3geneva2020openvins} is an open-source visual inertial navigation system developed by Geneva et al., which achieves localization accuracy comparable to that of SOTA optimization-based VIO on several public datasets. Several recent works aim to improve the robustness of MSCKF, with a particular focus on online calibration\cite{yang2023online}, observability consistency\cite{huang2025consistency}, and numerical stability\cite{wu2015square}.

Line features have been proven effective in detecting rich and semantically meaningful structures within challenging environments \cite{yao2024sg}. Building upon this advantage, a stereo visual-inertial navigation system that integrates both point and line features, Trifo-VIO \cite{zheng2018trifo}, has been developed. Yang et al. combined line features or plane features to improve the positioning accuracy of OpenVINS \cite{yang2019observability}. In \cite{10wei2021point}, SLAM line features were introduced for the first time, along with a prediction-matching line tracking strategy. To further improve the accuracy of line feature initialization, a two-step triangulation method was subsequently proposed \cite{chen2024fast}.

Almost all variants of MSCKF suffer from linearization errors and delayed measurement updates. On-manifold KF techniques have been employed to address the linearization errors in MSCKF, leading to notable improvements in both estimation consistency and accuracy\cite{barrau2016invariant}. Bloesch et al. introduced iterative EKF into MSCKF to improve the accuracy, but made the update more complicated\cite{bloesch2017iterated}. The equivariant MSCEqF algorithm \cite{fornasier2023msceqf} achieves consistent and reliable estimation even when the initial linearization points of calibration parameters deviate substantially from their true values. Compared to delayed updates, immediate updates incorporate more observation constraints and allow more frequent filtering updates, thus enhances the estimation accuracy\cite{zhang2025immediate}. However, the feature coordinates estimation errors still persist. The PO-KF\cite{9wang2025po} algorithm, which adopts a pose-only representation of point features, explicitly eliminates feature coordinates from the measurement equation. This system avoids linearization errors and enables immediate updates, which improves localization accuracy while maintaining real-time performance. SP-VIO \cite{du2024sp} introduces a double state transformation EKF and pose-only visual representation into the VIO framework, aiming to balance accuracy, efficiency, and robustness.

The pose-only representation has demonstrated strong potential in mitigating  linearization errors and delayed updates.  Based on this, we derive a pose-only geometric representation for line features and integrate pose-only representations of both point and line features into a filter-based VIO system.

\section{PRELIMINARIES}
\label{pre}
\subsection{Notations}
We define the notations used throughout this paper. $^{G}(\cdot)$, $^{I}(\cdot )$, and $^{C}(\cdot )$ denote the world frame, IMU frame (or body frame), and camera frame, respectively. $(\hat{\cdot })$ denotes the estimation state and $(\tilde{\cdot})$ represents the error state. ${}_{x}^{y}\mathbf{R}$ denotes the rotation matrix from $x$ frame to $y$ frame, and ${}_{\text{ }\!\!~\!\!\text{ }}^{y}{{\mathbf{p}}_{x}}$ represents the position of the origin of frame $x$ expressed in $y$ frame.
\subsection{Multi-State Constraint Kalman Filter}
\textit{1) State Definition:} The state vector is defined as
\begin{equation}
	\label{deqn_ex1}
	\mathbf{x}={{\left[ \begin{matrix}
				\mathbf{x}_{I}^{T} & \begin{matrix}
					\mathbf{x}_{cam}^{\text{T}} & \mathbf{x}_{ext}^{\text{T}}  \\
				\end{matrix} & \mathbf{x}_{C}^{T} & \mathbf{x}_{P}^{\text{T}} & \mathbf{x}_{L}^{\text{T}}  \\
			\end{matrix} \right]}^{\text{T}}},
\end{equation}
where ${{\mathbf{x}}_{I}}$ is the current IMU state, ${{\mathbf{x}}_{C}}$ denotes a sliding window of $W$ cloned IMU states, $\mathbf{x}_{cam}$ and $\mathbf{x}_{ext}^{{}}$ represent the camera intrinsic and extrinsic parameters. ${{\mathbf{x}}_{P}}$ and ${{\mathbf{x}}_{L}}$ are $M$ SLAM point and $N$ SLAM lines. The IMU state ${{\mathbf{x}}_{I}}$ is described as
\begin{equation}
	\label{deqn_ex2}
	{{\mathbf{x}}_{I}}={{\left[ \begin{matrix}
				_{G}^{I}{{{\mathbf{\bar{q}}}}^{\text{T}}} & ^{G}\mathbf{p}_{I}^{\text{T}} & ^{G}\mathbf{v}_{I}^{\text{T}} & \mathbf{b}_{g}^{\text{T}} & \mathbf{b}_{a}^{\text{T}}  \\
			\end{matrix} \right]}^{\text{T}}},
\end{equation}
where $_{G}^{I}{{\mathbf{\bar{q}}}^{\text{T}}}$ is the unit quaternion, ${}^{G}\mathbf{p}_{I}$ and ${}^{G}\mathbf{v}_{I}$ are position and velocity, and $\mathbf{b}_{g}$ and $\mathbf{b}_{a}$ are the gyroscope and accelerometer biases of IMU, respectively. $\mathbf{x}_{C}$ is defined as:
\begin{align}
	\label{deqn_ex3}
	\mathbf{x}_C &= \left[ 
	\begin{matrix}
		{}_{G}^{I_k}\bar{\mathbf{q}}^\mathrm{T} & 
		{}^{G}\mathbf{p}_{I_k}^\mathrm{T} & 
		{}_{G}^{I_{k-1}}\bar{\mathbf{q}}^\mathrm{T} & 
		{}^{G}\mathbf{p}_{I_{k-1}}^\mathrm{T}
	\end{matrix}
	\right. \nonumber \\
	& \quad\quad\quad \left.
	\begin{matrix}
		\cdots & 
		{}_{G}^{I_{k-W+1}}\bar{\mathbf{q}}^\mathrm{T} & 
		{}^{G}\mathbf{p}_{I_{k-W+1}}^\mathrm{T}
	\end{matrix}
	\right]^\mathrm{T}.
\end{align}

The state representations of point and line features have the following form:
\begin{align}
	{{\mathbf{x}}_{P}} &= 
	\begin{bmatrix}
		^{G}\mathbf{P}_{f1}^{T} & ^{G}\mathbf{P}_{f2}^{T} & \cdots & ^{G}\mathbf{P}_{fM}^{T}
	\end{bmatrix}^{\text{T}}, \label{deqn_ex4} \\
	{{\mathbf{x}}_{L}} &= 
	\begin{bmatrix}
		^{G}\mathbf{L}_{1}^{T} & ^{G}\mathbf{L}_{2}^{T} & \cdots & ^{G}\mathbf{L}_{N}^{T}
	\end{bmatrix}^{\text{T}}, \label{deqn_ex5}
\end{align}
where $^{G}\mathbf{P}_f$ is a $3 \times 1  $ vector, and ${}^{G}\mathbf{L}$ is a $4 \times 1  $ vector, represented using the closest point parameterization\cite{11yang2019aided}. The camera intrinsic parameters $\mathbf{x}_{cam}^{{}}$ and extrinsic parameters $\mathbf{x}_{ext}^{{}}$ are defined as 
\begin{align}
	\label{deqn_ex6}
	{{\mathbf{x}}_{cam}} &= 
	{{\left[ \begin{matrix}
				{{k}_{1}} & {{k}_{2}} & {{p}_{1}} & {{p}_{2}} & {{f}_{x}} & {{f}_{y}} & {{c}_{x}} & {{c}_{y}}  
			\end{matrix} \right]}^{\text{T}}}, \\
	\label{deqn_ex7}
	{{\mathbf{x}}_{ext}} &= 
	{{\left[ \begin{matrix}
				_{I}^{C}{{{\mathbf{\bar{q}}}}^{\text{T}}} & {}_{~}^{C}{{\mathbf{P}}_{I}}^{\text{T}}  
			\end{matrix} \right]}^{\text{T}}},
\end{align}
where $k_1$, $k_2$ and $p_1$, $p_2$  are the radial and tangential distortion coefficients,respectively. $f_x$ and $f_y$ represent the focal length, and $c_x$ and $c_y$ represent the offset between the camera optical center and the image center point on the pixel plane.

\textit{2) Propagation and Augmentation:} Since low-cost IMUs are insensitive to the Earth's rotation, the Earth's rotation can be neglected in their dynamic model. The angular velocity and acceleration measurement models are simplified as follows:
\begin{align}
	\label{deqn_ex8}
	\mathbf{\omega}_{m} &= \mathbf{\omega} + \mathbf{b}_{g} + \mathbf{n}_{g}, \\
	\label{deqn_ex9}
	\mathbf{a}_{m} &= \mathbf{a} + {}_{G}^{I}\mathbf{R} \, {}^{G}\mathbf{g} + \mathbf{b}_{a} + \mathbf{n}_{a},
\end{align}
where ${{\mathbf{\omega }}_{m}}$ and ${{\mathbf{a}}_{m}}$ denote the measured angular velocity and acceleration, respectively. $\mathbf{\omega }$ and $\mathbf{a}$ represent the true angular velocity and acceleration. ${{\mathbf{n}}_{g}} $ and ${{\mathbf{n}}_{a}}$ are the zero mean Gaussian white noise of gyroscope and accelerometer. The IMU kinematic model is formulated as
\begin{equation}
	\label{deqn_ex11}
	\begin{aligned}
	_{G}^{I}{{{\mathbf{\dot{\bar{q}}}}}^{I}} & =\frac{1}{2}\mathbf{\Omega }(\mathbf{\omega })_{G}^{I}\mathbf{\bar{q}},\text{ }{}^{\text{G}}{{{\mathbf{\dot{P}}}}_{I}}  ={}^{\text{G}}{{\mathbf{v}}_{I}}, {{\text{ }}^{G}}{{{\mathbf{\dot{v}}}}_{I}}  ={}_{G}^{I}{{\mathbf{R}}^{\text{T}}}\mathbf{a}, \\ 
	{{{\mathbf{\dot{b}}}}_{g}}  	& ={{\mathbf{n}}_{wg}}, {{{\mathbf{\dot{b}}}}_{a}}  ={{\mathbf{n}}_{wa}},  
	\end{aligned}
\end{equation}
where $\mathbf{\Omega }(\mathbf{\omega }) =\left[ \begin{array}{*{35}{l}}
	-{{\left[ \mathbf{\omega } \right]}_{\times }} & \mathbf{\omega }  \\
	-{{\mathbf{\omega }}^{\text{T}}} & 0  \\
\end{array} \right]$, ${\left[ \mathbf{\omega } \right]}_{\times }$ denotes the skew-symmetric matrix of $\mathbf{\omega }$.

\begin{figure*}[ht]
	\centering
	\includegraphics[width=0.86\linewidth]{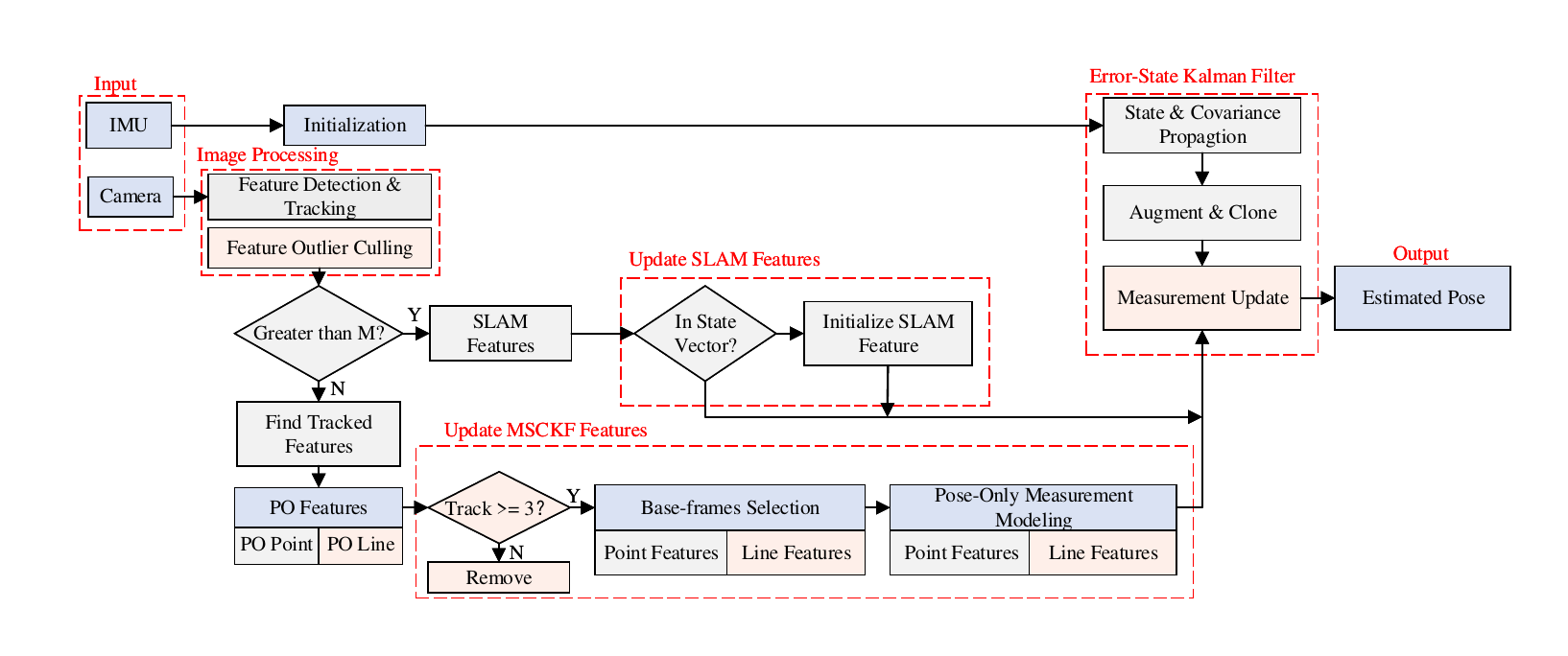}
	\caption{System overview of POPL-KF. The light pink-filled boxes represent the processing steps of pose-only line features.}
	\label{fig_1}
\end{figure*}
The other system states, including ${{\mathbf{x}}_{C}}$, ${{\mathbf{x}}_{P}}$,
${{\mathbf{x}}_{L}}$, $\mathbf{x}_{cam}^{}$ and $\mathbf{x}_{ext}^{}$ 
remain unchanged during the propagation phase. Consequently, their corresponding Jacobian matrices are unit matrices and the associated process noise is zero. By discretizing the IMU motion equations and stacking the above state transition equations, the complete system error state propagation model can be formulated as follows:
\begin{equation}
	\label{deqn_ex12}
	\mathbf{\tilde{x}}(k+1|k) =\mathbf{\Phi }(k+1|k)  \mathbf{\tilde{x}}(k+1|k)  +\mathbf{G}(k+1|k)  {{\mathbf{n}}_{k}},
\end{equation}
where $\mathbf{\Phi }(k+1|k)$ and $\mathbf{G}(k+1|k)$ represent the discrete-time state transition and noise-driven matrix, respectively, and ${{\mathbf{n}}_{k}}$ is the discrete noise vector with covariance matrix ${{\mathbf{Q}}_{{k}}}$. The covariance propagation is computed as:
\begin{align}
	\label{deqn_ex13}
	\mathbf{P}(k+1|k) =\; & \mathbf{\Phi}(k+1,k)\, \mathbf{P}(k|k)\, \mathbf{\Phi}^{\text{T}}(k+1,k) \nonumber \\
	& + \mathbf{G}(k+1,k)\, \mathbf{Q}_{k}\, \mathbf{G}^{\text{T}}(k+1,k).
\end{align}

The state and covariance augmentation procedures follow the implementation in OpenVINS\cite{3geneva2020openvins} and are omitted here for brevity.

\textit{3) Measurement update:}  When a point or line feature is lost in the current frame or reaches the maximum tracking length, feature coordinates are triangulated and then projected onto the image plane to  formulate the measurement equation. For a SLAM features, we first check whether the feature has been added to the state vector. If not, it must be initialized and then added to the state vector. Otherwise, its new measurement is directly used to update the state vector and covariance matrix. After stacking the measurements of all features, the complete measurement equation can be formulated as
\begin{equation}
	\label{deqn_ex13}
	\mathbf{z}= {{\mathbf{H}}_{\mathbf{x}}}\mathbf{\tilde{x}}+ {{\mathbf{H}}_{\mathbf{P}}}^{G}{{\mathbf{\tilde{P}}}_{f}}+{{\mathbf{H}}_{\mathbf{L}}}{}^{G}\mathbf{\tilde{L}}+{{\mathbf{n}}^{w}},
\end{equation}
where ${{\mathbf{H}}_{\mathbf{x}}}$, ${{\mathbf{H}}_{\mathbf{P}}}$ and ${{\mathbf{H}}_{\mathbf{L}}}$ denote the measurement Jacobian matrices with respect to the state vector, point features and line feature coordinates, respectively, ${{\mathbf{n}}^{w}}$ represents the measurement noise. Then, null-space projection is performed to eliminate the feature coordinates from \eqref{deqn_ex13}.

\section{Overview}
\label{ove}
Fig.~\ref{fig_1} illustrates the overall pipeline of the proposed POPL-KF system. The system is developed based on OpenVINS and includes four modules: image processing, initialization, state propagation, and measurement update. Our implementation retains the original design of OpenVINS in terms of initialization, IMU propagation, and point feature processing. On top of this foundation, we introduce the key enhancements: 1) we construct the measurement equations solely based on the poses of the base-frames, which reduces linearization errors and enables immediate updates. 2) A unified base-frame selection algorithm for both point and line features is designed to ensure optimal constraints on camera poses within the pose-only measurement model. 3) A visual front-end for line feature extraction and tracking is integrated into the POPL-KF system.
\section{VIO With Pose-Only representation of Point and Line Features}
\label{poseonly}
\subsection{Image Processing}
POPL-KF employs the same point feature detection and tracking algorithm as OpenVINS. For line features, we integrate an EDLines-based detector and a KLT-based tracking method. Given line features are more susceptible to introducing errors that may degrade system performance, their use must be more carefully managed. To enhance their robustness,  we propose a culling strategy based on grid-based image partitioning and bidirectional optical flow consistency. When the image contains abundant line features, EDLines tends to extract a large number of dense line segments, leading to a sharp increase in the number of tracked line features. This significantly raises the computational load for feature matching. Moreover, the short distance between line segments can lead to incorrect feature matches. First, to address the issue of excessively dense line features extracted by EDLines, the image is divided into multiple grid cells. If the number of line segments within a grid exceeds the preset threshold, only a subset of the longer line segments are retained, which reduces redundant line features and alleviates the challenges in line matching caused by local feature density. Then, we perform bidirectional optical flow on anchor points uniformly sampled along the line features to assess the consistency. Tracking is conducted both from the current frame to the previous frame and vice versa to evaluate flow consistency. If the majority of anchor points on a given line segment fail to maintain positional consistency during bidirectional tracking, the corresponding line feature is considered unreliable and subsequently discarded. This strategy enhances the overall observation quality and contributes to improved estimation accuracy within the system.

\subsection{Pose-only Measurement Model for Point Feature}

With reference to the pose-only theory\cite{7cai2021pose}, PO-KF\cite{9wang2025po} utilizes the pose of base-frames ${}^{G}{{\mathbf{T}}_{{{C}}}}=\left\{ {}_{G}^{{{C}}}\mathbf{R},\text{ }{}^{G}{{\mathbf{p}}_{{{C}}}} \right\}$ to construct the pose-only measurement equations. 

Considering a feature point $^{G}\mathbf{P}_f$ in the world frame is observed in $n$ image frames, its position in each camera frame can be derived using the camera poses as follows:
\begin{equation}
	\label{deqn_ex14}
	{}_{~}^{G}{{\mathbf{P}}_f}=  {}_{G}^{{{C}_{i}}}{{\mathbf{R}}^{T}}{}_{~}^{{{C}_{i}}}{{\mathbf{P}}_{f}}+{}_{~}^{G}{{\mathbf{P}}_{{{C}_{i}}}}=  {}_{~}^{{{C}_{i}}}z{}_{G}^{{{C}_{i}}}{{\mathbf{R}}^{T}}{}_{~}^{{{C}_{i}}}\mathbf{f}+{}_{~}^{G}{{\mathbf{P}}_{{{C}_{i}}}},
\end{equation}
where ${}_{~}^{{{C}_{i}}}\mathbf{f}$ represents the normalized plane coordinates of the feature point in the ${{C}_{i}}$ frame, ${}_{~}^{{{C}_{i}}}z$ is the feature depth in the ${{C}_{i}}$ frame.

Let $\left( i,j \right)$ be a pair of views consisting of ${{C}_{i}}$ and ${{C}_{j}}$ images. The projection relationship between the two views can be expressed as:
\begin{equation}
	\label{deqn_ex15}
     {}_{~}^{{{C}_{j}}}z{}_{~}^{{{C}_{j}}}\mathbf{f}={}_{~}^{{{C}_{i}}}z{}_{{{C}_{i}}}^{{{C}_{j}}}\mathbf{R}{}_{~}^{{{C}_{i}}}\mathbf{f}+{}_{~}^{{{C}_{j}}}{{\mathbf{P}}_{{{C}_{i}}}},
\end{equation}
where ${}_{~}^{{{C}_{j}}}{{\mathbf{P}}_{{{C}_{i}}}}  ={}_{G}^{{{C}_{j}}}\mathbf{R}\left( {}_{~}^{G}{{\mathbf{P}}_{{{C}_{i}}}}-{}_{~}^{G}{{\mathbf{P}}_{{{C}_{j}}}} \right)$,    ${}_{{{C}_{i}}}^{{{C}_{j}}}\mathbf{R}  ={}_{G}^{{{C}_{j}}}\mathbf{R}{}_{G}^{{{C}_{i}}}{{\mathbf{R}}^{T}}$.

Since both ${}_{~}^{{{C}_{i}}}z$ and ${}_{~}^{{{C}_{j}}}z$ are unknown, we multiply both sides of \eqref{deqn_ex15} by ${{\left[ {}_{~}^{{{C}_{j}}}\mathbf{f} \right]}_{\times }}$,  yielding
\begin{equation}
	\label{deqn_ex16}
	{}_{~}^{{{C}_{i}}}z{{\left[ {}_{~}^{{{C}_{j}}}\mathbf{f} \right]}_{\times }}{}_{{{C}_{i}}}^{{{C}_{j}}}\mathbf{R}{}_{~}^{{{C}_{i}}}\mathbf{f}=-{{\left[ {}_{~}^{{{C}_{j}}}\mathbf{f} \right]}_{\times }}{}_{~}^{{{C}_{j}}}{{\mathbf{P}}_{{{C}_{i}}}}.
\end{equation}

Taking the magnitude of \eqref{deqn_ex16}, the depth ${}_{~}^{{{C}_{i}}}z$ can be computed as
\begin{equation}
	\label{deqn_ex17}
	{}_{~}^{{{C}_{i}}}z= \frac{\left\| -{{\left[ {}_{~}^{{{C}_{j}}}\mathbf{f} \right]}_{\times }}{}_{~}^{{{C}_{j}}}{{\mathbf{P}}_{{{C}_{i}}}} \right\|}{\left\| {{\left[ {}_{~}^{{{C}_{j}}}\mathbf{f} \right]}_{\times }}{}_{{{C}_{i}}}^{{{C}_{j}}}\mathbf{R}{}_{~}^{{{C}_{i}}}\mathbf{f} \right\|}={{\text{h}}_{{}_{~}^{{{C}_{i}}}z}}\left( {}^{G}{{\mathbf{T}}_{{{C}_{i}}}},\text{ }{}^{G}{{\mathbf{T}}_{{{C}_{j}}}} \right)=z_{i}^{\left( i,j \right)},
\end{equation}
where $z_{i}^{(i,j)}$ denotes the depth of the feature, constrained by the image pair $(i,j)$ in the ${{C}_{i}}$ frame.

Similarly, the depth in the ${{C}_{j}}$ frame is denoted as $z_{j}^{(i,j)}$. The feature coordinates can be expressed in terms of the camera pose and normalized observation as
\begin{equation}
	\label{deqn_ex18}
	{}_{~}^{{{C}_{i}}}{{\mathbf{P}}_{f}}=  z_{i}^{\left( i,j \right)}{}_{~}^{{{C}_{i}}}\mathbf{f},{}_{~}^{{{C}_{j}}}{{\mathbf{P}}_{f}}=  z_{j}^{\left( i,j \right)}{}_{~}^{{{C}_{j}}}\mathbf{f}.
\end{equation}

When a point feature is observed in the ${{C}_{k}}$ frame, we construct the measurement equation using the current observation $\mathbf{z}_{k}^{{}}$ and ${}_{~}^{G}{{\mathbf{P}_f}}$ obtained from the image pair $\left( i,j \right)$ , where ${}_{~}^{G}\mathbf{P}_f ={{\text{h}}_{{}_{~}^{G}\mathbf{P}}}\left( {}^{G}{{\mathbf{T}}_{{{C}_{i}}}},{}_{~}^{{{C}_{i}}}z \right)$, as defined in \eqref{deqn_ex14}. The coordinates of the feature point in the ${{C}_{k}}$ frame are then expressed as
\begin{equation}
	\label{deqn_ex19}
	{}_{~}^{{{C}_{k}}}{{\mathbf{P}}_{f}}\!=\!{}_{G}^{{{C}_{k}}}\mathbf{R}\left( {}_{~}^{G}{{\mathbf{P}}_f	}\!-\!{}_{~}^{G}{{\mathbf{P}}_{{{C}_{k}}}} \right)\!=\!{{\text{h}}_{t}}\left( {}_{~}^{G}{{\mathbf{P}}_f},{}^{G}{{\mathbf{T}}_{{{C}_{k}}}} \right).
\end{equation}

The normalized image plane coordinates of the feature point in the ${{C}_{k}}$ frame are estimated as
\begin{equation}
	\label{deqn_ex20}
	{}_{~}^{{{C}_{k}}}\mathbf{\hat{f}}=\frac{{}_{~}^{{{C}_{k}}}{{\mathbf{P}}_{f}}}{\left[ \begin{matrix}
			0 & 0 & 1  \\
		\end{matrix} \right]{}_{~}^{{{C}_{k}}}{{\mathbf{P}}_{f}}}=\left[ \begin{matrix}
		{{u}_{k}} & {{v}_{k}} & 1  \\
	\end{matrix} \right]\text{=}{{\text{h}}_{u}}\left( {}_{~}^{{{C}_{k}}}{{\mathbf{P}}_{f}} \right).
\end{equation}

Radial-tangential distortion is applied to the normalized coordinates, which are then projected onto the image plane. The pixel coordinates can be expressed as
\begin{equation}
	\label{deqn_ex21}
	{{\mathbf{\hat{z}}}_{k}} = {{\operatorname{h}}_{p}}{{\text{h}}_{d}}\left( \left( {}^{C_{k}}\mathbf{\hat{f}}, {{\mathbf{x}}_{cam}} \right), {{\mathbf{x}}_{cam}} \right),
\end{equation}
where ${\text{h}_{p}}$ and ${{\text{h}}_{d}}$ are the projection and distortion functions, which are detailed in\cite{9wang2025po}. The measurement equation for point features is defined as follows:
\begin{equation}
	\label{deqn_ex22}
	\begin{aligned}
		\mathbf{e}_{k}^{\mathbf{P}} &= \mathbf{z}_{k} - \hat{\mathbf{z}}_{k} + \mathbf{n}_{f,k} \\
		&= \mathbf{z}_{k} - \text{h}_{p} \Big( 
		\text{h}_{d} \Big( 
		\text{h}_{u} \Big( 
		\text{h}_{t} ( 
		{}^{G}\mathbf{T}_{C_i},\ 
		{}^{G}\mathbf{T}_{C_k}, \\
		&\quad\quad {\text{h}_{{}_{~}^{{{C}_{i}}}z}} (
		{}^{G}\mathbf{T}_{C_i},\ 
		{}^{G}\mathbf{T}_{C_j} ) 
		) 
		\Big),\ 
		\mathbf{x}_{cam} 
		\Big),\ 
		\mathbf{x}_{cam} 
		\Big) + \mathbf{n}_{f,k},
	\end{aligned}
\end{equation}
where ${{\mathbf{n}}_{f,k}}$ denotes the point measurement noise. The goal of the VIO system is to estimate the pose in the IMU frame ${}^{G}{{\mathbf{T}}_{I}} = \left\{ {}_{G}^{I}\mathbf{R},\ {}^{G}{{\mathbf{p}}_{I}} \right\}$, the transformation ${}^{G}{{\mathbf{T}}_{I}}$ and ${}^{G}{{\mathbf{T}}_{C}}$ is given by 
\begin{equation}
	\label{deqn_ex23}
	\begin{aligned}
		 {}_{G}^{{{C}_{\eta }}}\mathbf{R} &= {}_{I}^{C}\mathbf{R}{}_{G}^{{{I}_{\eta }}}\mathbf{R}, \\ 
		 {}^{G}{{\mathbf{p}}_{{{C}_{\eta }}}} &=  {}^{G}{{\mathbf{p}}_{{{I}_{\eta }}}}-{}_{G}^{{{C}_{\eta }}}{{\mathbf{R}}^{\text{T}}}{}_{~}^{C}{{\mathbf{p}}_{I}}, \\ 
		 \Rightarrow {}^{G}{{\mathbf{T}}_{{{C}_{\eta }}}} &= {{\text{h}}_{{}^{G}{{\mathbf{T}}_{{{C}_{\eta }}}}}}\left( {}^{G}{{\mathbf{T}}_{{{I}_{\eta }}}},\text{ }{{\mathbf{x}}_{ext}} \right),\text{ }\eta =i,\text{ }j,\text{ }k.
	\end{aligned}
\end{equation}

It can be observed that the pose-only point features measurement equation $\mathbf{e}_{k}^{\mathbf{P}}$ depends solely on the IMU poses of the base frames, as well as the camera intrinsics and extrinsics. Therefore, \eqref{deqn_ex23} can be simplified as
\begin{equation}
	\label{deqn_ex24}
	\begin{aligned}
		\mathbf{e}_{k}^{\mathbf{P}}=&{{\mathbf{H}}_{{{\mathbf{T}}_{{{I}_{i}}}}}}{}_{~}^{G}{{\mathbf{\tilde{T}}}_{{{I}_{i}}}}+{{\mathbf{H}}_{{{\mathbf{T}}_{{{I}_{j}}}}}}{}^{G}{{\mathbf{\tilde{T}}}_{{{I}_{j}}}}+{{\mathbf{H}}_{{{\mathbf{T}}_{{{I}_{k}}}}}}{}_{~}^{G}{{\mathbf{\tilde{T}}}_{{{I}_{k}}}}\\&+{{\mathbf{H}}_{cam}}{{\mathbf{\tilde{x}}}_{cam}}+{{\mathbf{H}}_{ext}}{{\mathbf{\tilde{x}}}_{ext}}+{{\mathbf{n}}_{f,k}},
	\end{aligned}
\end{equation}
where ${{\mathbf{H}}_{{{\mathbf{T}}_{{{I}_{i}}}}}}$ is the Jacobian matrix with respect to the IMU pose, ${{\mathbf{H}}_{cam}}$ and ${{\mathbf{H}}_{ext}}$ represent the Jacobian matrices of the camera intrinsic and extrinsic parameters, respectively. The complete expressions of the Jacobian matrices are provided in Appendix A.

\subsection{Pose-only Measurement Model for Line Feature}

Inspired by the effectiveness of pose-only representations for point features, we propose a pose-only geometric representation for line features, derived from classical multi-view geometry. This representation enables efficient camera motion estimation and provides clear physical interpretability. In this subsection, we present the mathematical derivation and the associated measurement equations.

\begin{figure}[ht]
	\centering
	\includegraphics[width=\linewidth]{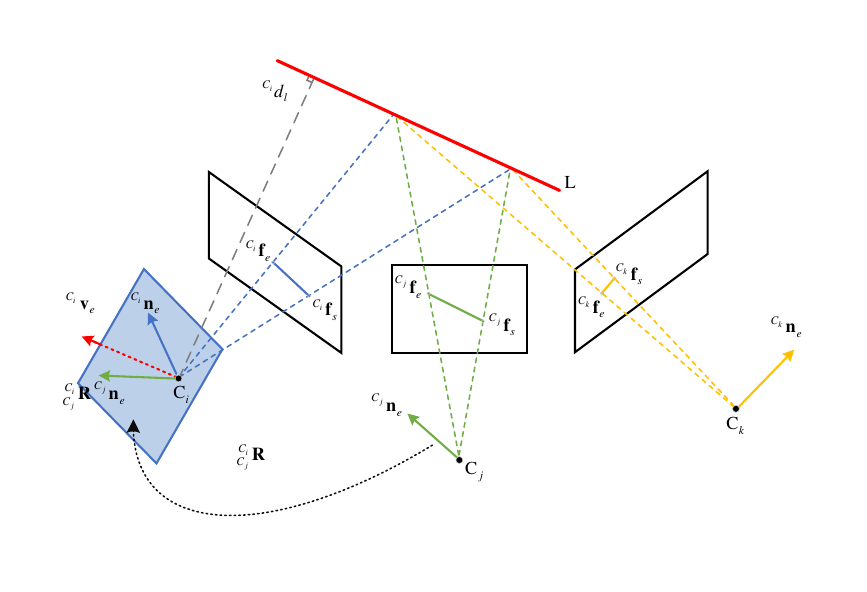}
	\caption{Illustration of pose-only representation for line features.}
	\label{fig_2}
\end{figure}

\textit{1) Direction vector pose-only constraint:}  Consider a spatial line feature ${}^{G}\mathbf{L}$ observed in $n$ historical frames. For each frame $i = 1, 2, \ldots, n$, the unit normal vector of the corresponding projection plane is defined as
\begin{equation}
	\label{deqn_ex24ad2}
	{}^{C_{i}}{{\mathbf{n}}_{e}}=\frac{{{\left[ {}^{C_{i}}{{\mathbf{f}}_{s}} \right]}_{\times }}{}^{C_{i}}{{\mathbf{f}}_{e}}}{\left\| {{\left[ {}^{C_{i}}{{\mathbf{f}}_{s}} \right]}_{\times }}{}^{C_{i}}{{\mathbf{f}}_{e}} \right\|},
\end{equation}
where ${}^{C_{i}}\mathbf{f}_{s}$ and ${}^{C_{i}}\mathbf{f}_{e}$ denote the normalized homogeneous coordinates of the start and end points of the line in the $C_{i}$ frame, respectively.

The projection plane normal vectors of line features observed in different frames are orthogonal to the spatial line direction vector. Therefore, the normal vectors ${}^{C_{i}}{{\mathbf{n}}_{e}}$ and the transformed normal vector ${}_{C_{j}}^{C_{i}}\mathbf{R}{}^{C_{j}}{{\mathbf{n}}_{e}}$ lie in the same plane in the $C_{i}$ frame, as shown in Fig.~\ref{fig_2}. The direction vector of the line in the $C_{i}$ frame, ${}^{{C}_{i}}{{\mathbf{v}}_{e}} $, is aligned with the normal vector of this plane, as formulated:
\begin{equation}
	\label{deqn_ex25}
	{}^{{C}_{i}}{{\mathbf{v}}_{e}} = \operatorname{sgn}(\zeta) \frac{{\left[ {}_{{C}_{j}}^{{C}_{i}}\mathbf{R} \, {}^{{C}_{j}}{{\mathbf{n}}_{e}} \right]_{\times}} \, {}^{{C}_{i}}{{\mathbf{n}}_{e}}}{\left\| {\left[ {}_{{C}_{j}}^{{C}_{i}}\mathbf{R} \, {}^{{C}_{j}}{{\mathbf{n}}_{e}} \right]_{\times}} \, {}^{{C}_{i}}{{\mathbf{n}}_{e}} \right\|} = \mathbf{v}_{i}^{(i,j)},
\end{equation}
where $\mathbf{v}_{i}^{(i,j)}$ represents the direction vector pose-only constraint in the $C_{i}$ frame, constrained by the image pair $\left(i, j\right)$, and $\operatorname{sgn}(\cdot)$ is the sign function. To ensure numerical consistency with the directional constraint, we define $\zeta$ as follows:
\begin{equation}
	\label{deqn_ex26}
	\zeta ={{({}^{{{C}_{i}}}{{\mathbf{f}}_{e}}-{}^{{{C}_{i}}}{{\mathbf{f}}_{s}})}^{\text{T}}}{{\left[ {}_{{{C}_{j}}}^{{{C}_{i}}}\mathbf{R}{}^{{{C}_{j}}}{{\mathbf{n}}_{e}} \right]}_{\times }}{}^{{{C}_{i}}}{{\mathbf{n}}_{e}},
\end{equation}

Similarly, the direction vector constrained by the image pair $\left(i, j\right)$ in the $C_{j}$ frame can be expressed as
\begin{equation}
	\label{deqn_ex27}
	{}^{{{C}_{j}}}{{\mathbf{v}}_{e}}=\operatorname{sgn}(\zeta )\frac{{{\left[ {}_{{{C}_{i}}}^{{{C}_{j}}}\mathbf{R}{}^{{{C}_{i}}}{{\mathbf{n}}_{e}} \right]}_{\times }}{}^{{{C}_{j}}}{{\mathbf{n}}_{e}}}{\left\| {{\left[ {}_{{{C}_{i}}}^{{{C}_{j}}}\mathbf{R}{}^{{{C}_{i}}}{{\mathbf{n}}_{e}} \right]}_{\times }}{}^{{{C}_{j}}}{{\mathbf{n}}_{e}} \right\|}=\mathbf{v}_{j}^{(i,j)}.
\end{equation}

\textit{2) Distance pose-only constraint:} The coordinate transformation for line features is given by
\begin{equation}
	\label{deqn_ex28}
	\left[ \begin{matrix}
		{}^{{{C}_{i}}}{{d}_{l}}{}^{{{C}_{i}}}{{\mathbf{n}}_{e}}  \\
		^{{{C}_{i}}}{{\mathbf{v}}_{e}}  \\
	\end{matrix} \right]=\left[ \begin{matrix}
		{}_{{{C}_{j}}}^{{{C}_{i}}}\mathbf{R} & {{\left[ {}^{{{C}_{i}}}{{\mathbf{P}}_{{{C}_{j}}}} \right]}_{\times }}{}_{{{C}_{j}}}^{{{C}_{i}}}\mathbf{R}  \\
		{{\mathbf{O}}_{3\times 3}} & {}_{{{C}_{j}}}^{{{C}_{i}}}\mathbf{R}  \\
	\end{matrix} \right]\left[ \begin{matrix}
	{}^{{{C}_{j}}}{{d}_{l}}{}^{{{C}_{j}}}{{\mathbf{n}}_{e}}  \\
		^{{{C}_{j}}}{{\mathbf{v}}_{e}}  \\
	\end{matrix} \right],
\end{equation}
where ${}^{C_{i}}d_{l}$ and ${}^{C_{j}}d_{l}$ represent the distances from the origins of $C_{i}$ and $C_{j}$ to the spatial line, respectively. Then we have
\begin{equation}
	\label{deqn_ex29}
	{}^{{{C}_{i}}}{{d}_{l}}{}^{{{C}_{i}}}{{\mathbf{n}}_{e}}={}^{{{C}_{j}}}{{d}_{l}}{}_{{{C}_{j}}}^{{{C}_{i}}}\mathbf{R}{}^{{{C}_{j}}}{{\mathbf{n}}_{e}}+{{\left[ {}^{{{C}_{i}}}{{\mathbf{P}}_{{{C}_{j}}}} \right]}_{\times }}{}^{{{C}_{i}}}{{\mathbf{v}}_{e}}.
\end{equation}

Since both ${}^{C_{i}}d_{l}$ and ${}^{C_{j}}d_{l}$ are unknown, we multiply both sides of \eqref{deqn_ex15} by ${\left[ {}_{C_{j}}^{C_{i}}\mathbf{R}\,{}^{C_{j}}\mathbf{n}_{e} \right]}_{\times}$, yielding
\begin{equation}
	\label{deqn_ex30}
	{}^{{{C}_{i}}}{{d}_{l}}{{\left[ {}_{{{C}_{j}}}^{{{C}_{i}}}\mathbf{R}{}^{{{C}_{j}}}{{\mathbf{n}}_{e}} \right]}_{\times }}{}^{{{C}_{i}}}{{\mathbf{n}}_{e}}={{\left[ {}_{{{C}_{j}}}^{{{C}_{i}}}\mathbf{R}{}^{{{C}_{j}}}{{\mathbf{n}}_{e}} \right]}_{\times }}{{\left[ {}^{{{C}_{i}}}{{\mathbf{P}}_{{{C}_{j}}}} \right]}_{\times }}{}^{{{C}_{i}}}{{\mathbf{v}}_{e}}.
\end{equation}

Substituting \eqref{deqn_ex25} into \eqref{deqn_ex30} gives
\begin{equation}
	\label{deqn_ex31}
	\begin{aligned}
		{}^{{C}_{i}} d_{l} \left[ 
		{}_{{C}_{j}}^{{C}_{i}} \mathbf{R} \, {}^{{C}_{j}} \mathbf{n}_{e} 
		\right]_{\times} \, {}^{{C}_{i}} \mathbf{n}_{e}
		\!=\! &\operatorname{sgn}(\zeta) \left[ 
		{}_{{C}_{j}}^{{C}_{i}} \mathbf{R} \, {}^{{C}_{j}} \mathbf{n}_{e} 
		\right]_{\times} \left[ 
		{}^{{C}_{i}} \mathbf{P}_{{C}_{j}} 
		\right]_{\times} \\
		&\quad \cdot 
		\frac{
			\left[ 
			{}_{{C}_{j}}^{{C}_{i}} \mathbf{R} \, {}^{{C}_{j}} \mathbf{n}_{e} 
			\right]_{\times} \, {}^{{C}_{i}} \mathbf{n}_{e}
		}{
			\left\| 
			\left[ 
			{}_{{C}_{j}}^{{C}_{i}} \mathbf{R} \, {}^{{C}_{j}} \mathbf{n}_{e} 
			\right]_{\times} \, {}^{{C}_{i}} \mathbf{n}_{e}
			\right\|
		}.
	\end{aligned}
\end{equation}

Equation \eqref{deqn_ex31} contains a denominator term that complicates subsequent calculations. Additionally, since the direction vector lacks scale information, we introduce intermediate variables ${}^{C_{i}}\mathbf{v}_{\alpha}$ and ${}^{C_{i}}d_{l\alpha}$ to simplify \eqref{deqn_ex31}:
\begin{equation}
	\label{deqn_ex32}
	\begin{aligned}
		 {}^{{{C}_{i}}}{{\mathbf{v}}_{\alpha }}&=\left\| {{\left[ {}_{{{C}_{j}}}^{{{C}_{i}}}\mathbf{R}{}^{{{C}_{j}}}{{\mathbf{n}}_{e}} \right]}_{\times }}{}^{{{C}_{i}}}{{\mathbf{n}}_{e}} \right\|{}^{{{C}_{i}}}{{\mathbf{v}}_{e}} =  \operatorname{sgn}(\zeta ){{\left[ {}_{{{C}_{j}}}^{{{C}_{i}}}\mathbf{R}{}^{{{C}_{j}}}{{\mathbf{n}}_{e}} \right]}_{\times }}{}^{{{C}_{i}}}{{\mathbf{n}}_{e}}\\
		 &={{\text{h}}_{{}^{{{C}_{i}}}{{\mathbf{v}}_{\alpha }}}}\left( {}_{~}^{G}{{\mathbf{T}}_{{{C}_{i}}}},\text{ }{}_{~}^{G}{{\mathbf{T}}_{{{C}_{j}}}} \right) \\ 
		 {}^{{{C}_{i}}}{{d}_{l\alpha }} & =\left\| {{\left[ {}_{{{C}_{j}}}^{{{C}_{i}}}\mathbf{R}{}^{{{C}_{j}}}{{\mathbf{n}}_{e}} \right]}_{\times }}{}^{{{C}_{i}}}{{\mathbf{n}}_{e}} \right\|{}^{{{C}_{i}}}{{d}_{l}}.
	\end{aligned}
\end{equation}

Equation \eqref{deqn_ex31} can then be expressed as
\begin{equation}
	\label{deqn_ex33}
	\operatorname{sgn} (\zeta ){}^{{{C}_{i}}}{{d}_{l\alpha }}{}^{{{C}_{i}}}{{\mathbf{v}}_{\alpha }}={{\left[ {}_{{{C}_{j}}}^{{{C}_{i}}}\mathbf{R}{}^{{{C}_{j}}}{{\mathbf{n}}_{e}} \right]}_{\times }}{{\left[ {}^{{{C}_{i}}}{{\mathbf{P}}_{{{C}_{j}}}} \right]}_{\times }}{}^{{{C}_{i}}}{{\mathbf{v}}_{\alpha }}.
\end{equation}

Taking the magnitude of \eqref{deqn_ex33}, ${}^{C_{i}}d_{l\alpha}$ can be represented as
\begin{equation}
	\label{deqn_ex34}
	\begin{aligned}
		{}^{{C}_{i}} d_{l\alpha} 
		&= \frac{
			\left\| 
			\left[ {}_{{C}_{j}}^{{C}_{i}} \mathbf{R} \, {}^{{C}_{j}} \mathbf{n}_{e} \right]_{\times} 
			\left[ {}^{{C}_{i}} \mathbf{P}_{{C}_{j}} \right]_{\times} 
			{}^{{C}_{i}} \mathbf{v}_{\alpha}
			\right\|
		}{
			\left\| {}^{{C}_{i}} \mathbf{v}_{\alpha} \right\|
		} \\
		&= \text{h}_{{}^{{C}_{i}} d_{l\alpha}} 
		\left( {}^{G} \mathbf{T}_{{C}_{i}}, \, {}^{G} \mathbf{T}_{{C}_{j}} \right).
	\end{aligned}
\end{equation}

The distance ${}^{C_{i}}d_{l}$ is obtained as
\begin{equation}
	\label{deqn_ex35}
	{}^{{{C}_{i}}}{{d}_{l}}=\frac{\left\| {{\left[ {}_{{{C}_{j}}}^{{{C}_{i}}}\mathbf{R}{}^{{{C}_{j}}}{{\mathbf{n}}_{e}} \right]}_{\times }}{{\left[ {}^{{{C}_{i}}}{{\mathbf{P}}_{{{C}_{j}}}} \right]}_{\times }}{}^{{{C}_{i}}}{{\mathbf{v}}_{\alpha }} \right\|}{\left\| {}^{{{C}_{i}}}{{\mathbf{v}}_{\alpha }} \right\|\left\| {{\left[ {}_{{{C}_{j}}}^{{{C}_{i}}}\mathbf{R}{}^{{{C}_{j}}}{{\mathbf{n}}_{e}} \right]}_{\times }}{}^{{{C}_{i}}}{{\mathbf{n}}_{e}} \right\|}=d_{i}^{(i,j)},
\end{equation}
where $d_{i}^{(i,j)}$ denotes the distance constrained by the image pair $(i,j)$ in the $C_{i}$ frame. Similarly, the distance constrained by the image pair $(i,j)$ in the $C_{j}$ frame can be formulated as $d_{j}^{(i,j)}$.

Combining \eqref{deqn_ex27}, \eqref{deqn_ex28} and \eqref{deqn_ex35}, the pose-only constraint for the two-view imaging geometry, referred to as a pair of pose-only constraints is obtained as
\begin{equation}
	\label{deqn_ex36}
	d_{i}^{(i,j)}{}^{{{C}_{i}}}{{\mathbf{n}}_{e}}=d_{j}^{(i,j)}{}_{{{C}_{j}}}^{{{C}_{i}}}\mathbf{R}{}^{{{C}_{j}}}{{\mathbf{n}}_{e}}+{{\left[ {}^{{{C}_{i}}}{{\mathbf{P}}_{{{C}_{j}}}} \right]}_{\times }}\mathbf{v}_{i}^{(i,j)}.
\end{equation}

Regarding the $k$-th$( k \neq j) $ image, the view pair $(i,k)$ also satisfies the pose-only constraint:
\begin{equation}
	\label{deqn_ex36}
	d_{i}^{(i,k)}{}^{{{C}_{i}}}{{\mathbf{n}}_{e}}=d_{k}^{(i,k)}{}_{{{C}_{k}}}^{{{C}_{i}}}\mathbf{R}{}^{{{C}_{k}}}{{\mathbf{n}}_{e}}+{{\left[ {}^{{{C}_{i}}}{{\mathbf{P}}_{{{C}_{k}}}} \right]}_{\times }}\mathbf{v}_{i}^{(i,k)}.
\end{equation}
and direction vector pose-only constraints:
\begin{equation}
	\label{deqn_ex38}
	\mathbf{v}_{i}^{(i,j)}=\mathbf{v}_{i}^{(i,k)}={}^{{{C}_{i}}}{{\mathbf{v}}_{e}},
\end{equation}
and distance pose-only constraints:
\begin{equation}
	\label{deqn_ex39}
	d_{i}^{(i,j)}=d_{i}^{(i,k)}={}^{{{C}_{i}}}d_l.
\end{equation}

\textit{3) Measurement:} When a line feature is observed in the $C_k$ frame, we construct the measurement equation using this frame along with the distance and direction vector derived from the image pair $(i,j)$. By incorporating  \eqref{deqn_ex28} and substituting $^{C_i}d_{l\alpha}$, $^{C_i}\mathbf{n}_e$ and $^{C_i}\mathbf{v}_\alpha$, the normal direction of the line in the $C_k$ frame can be expressed as
\begin{equation}
	\label{deqn_ex40}
	\begin{aligned}
		{}^{C_k}\mathbf{n}_l 
		&= \left\| \left[ {}_{C_j}^{C_i} \mathbf{R} {}^{C_j} \mathbf{n}_e \right]_{\times} {}^{C_i} \mathbf{n}_e \right\| {}^{C_k} d_l {}^{C_k} \mathbf{n}_e \\
		&= {}^{C_i} d_{l\alpha} {}_{C_i}^{C_k} \mathbf{R} {}^{C_i} \mathbf{n}_e + \left[ {}^{C_k} \mathbf{P}_{C_i} \right]_{\times} {}_{C_i}^{C_k} \mathbf{R} {}^{C_i} \mathbf{v}_{\alpha} \\
		&= h_{{}^{C_k} \mathbf{n}_l} \left( {}_{~}^G \mathbf{T}_{C_i}, {}_{~}^G \mathbf{T}_{C_k}, {}^{C_i} \mathbf{v}_\alpha, {}^{C_i} d_{l\alpha} \right).
	\end{aligned}
\end{equation}

The projection and distortion functions are applied to project ${}^{{{C}_{k}}}{{\mathbf{n}}_{l}}$ onto the image plane:
\begin{equation}
	\label{deqn_ex41}
	{}^{{{C}_{k}}}{{\mathbf{I}}_{l}} \!=\!{{\text{h}}_{{}^{{{C}_{k}}}{{\mathbf{I}}_{k}}}}\!\left( {}^{{{C}_{k}}}{{\mathbf{n}}_{l}},{{\mathbf{x}}_{cam}}\right)\!\!=\!{{\left[ \begin{matrix}
				{{}^{{{C}_{k}}}{I}_{k1}} & {{}^{{{C}_{k}}}{I}_{k}}_{2} & {{}^{{{C}_{k}}}{I}_{k}}_{3}  \\
			\end{matrix} \right]}^{\text{T}}}.
\end{equation}

The distance between the projected line and endpoints ${}^{k}\mathbf{\bar{s}}$ and ${}^{k}\mathbf{\bar{e}}$ of the tracked line can be expressed as
\begin{equation}
	\label{deqn_ex22}
	\begin{aligned}
		{{d}_{k}}&={{\left[ \begin{matrix}
					\frac{{}^{k}{{{\mathbf{\bar{s}}}}^{T}}{}^{{{C}_{k}}}{{\mathbf{I}}_{k}}}{\sqrt{{}^{{{C}_{k}}}{{I}_{k1}}^{2}+{}^{{{C}_{k}}}{{I}_{k2}}^{2}}} & \frac{{}^{k}{{{\mathbf{\bar{e}}}}^{T}}{}^{{{C}_{k}}}{{\mathbf{I}}_{k}}}{\sqrt{{}^{{{C}_{k}}}{{I}_{k1}}^{2}+{}^{{{C}_{k}}}{{I}_{k2}}^{2}}}  \\
				\end{matrix} \right]}^{T}} \\
		&={{\text{h}}_{{{d}_{k}}}}\left( {}^{{{C}_{k}}}{{\mathbf{I}}_{k}},{{\mathbf{x}}_{cam}} \right),
	\end{aligned}
\end{equation}
where $^{k}\mathbf{\bar{s}} = \begin{bmatrix} u_{s} & v_{s} & 1 \end{bmatrix}^{\mathrm{T}}$ and $^{k}\mathbf{\bar{e}} = \begin{bmatrix} u_{e} & v_{e} & 1 \end{bmatrix}^{\mathrm{T}}$,  with $u_{s}$, $v_{s}$ and $u_{e}$, $v_{e}$ representing  the pixel coordinates of the start and end points, respectively. The measurement equation for a line feature is defined as follows:
\begin{equation}
	\label{deqn_ex42}
	\begin{aligned}
		\mathbf{e}_{k}^{\mathbf{L}} &= \mathbf{0}_{2 \times 1} - \hat{d}_k \\
		&= 
		\text{h}_{d_k} \Big( 
		\text{h}_{{}^{C_k}\mathbf{I}_k} \big( 
		\text{h}_{{}^{C_k} \mathbf{n}_l} \big( 
		\text{h}_{{}^{C_i} d_{l\alpha}} 
		( {}^{G}\mathbf{T}_{C_i},\; {}^{G}\mathbf{T}_{C_j} ), \\
		&\qquad 
		\text{h}_{{}^{C_i} \mathbf{v}_\alpha} 
		( {}^{G}\mathbf{T}_{C_i},\; {}^{G}\mathbf{T}_{C_j} ),\; 
		{}^{G}\mathbf{T}_{C_k},\; 
		{}^{G}\mathbf{T}_{C_i} 
		\big),\; 
		\mathbf{x}_{\text{cam}} 
		\big),\; 
		\mathbf{x}_{\text{cam}} 
		\Big).
	\end{aligned}
\end{equation}

According to \eqref{deqn_ex42}, the pose-only line features measurement equation $\mathbf{e}_{k}^{\mathbf{L}}$ depends solely on the IMU pose of the base frame, as well as the camera intrinsics and extrinsics. Linearizing $\mathbf{e}_{k}^{\mathbf{L}}$ yields
\begin{equation}
	\label{deqn_ex43}
	\begin{aligned}
		\mathbf{e}_k^{\mathbf{L}} = & 
		\mathbf{H}_{\mathbf{T}_{I_i}}^{\mathbf{L}}{}^{G} \tilde{\mathbf{T}}_{I_i}
		+ \mathbf{H}_{\mathbf{T}_{I_j}}^{\mathbf{L}}  {}^{G} \tilde{\mathbf{T}}_{I_j} \\
		& + \mathbf{H}_{\mathbf{T}_{I_k}}^{\mathbf{L}} {}^{G} \tilde{\mathbf{T}}_{I_k}
		+ \mathbf{H}_{\text{cam}}^{\mathbf{L}}  \tilde{\mathbf{x}}_{\text{cam}} 
		 + \mathbf{H}_{\text{ext}}^{\mathbf{L}}  \tilde{\mathbf{x}}_{\text{ext}}
		+ \mathbf{n}_{l,k},
	\end{aligned}
\end{equation}
where ${{\mathbf{n}}_{l,k}}$ denotes the line measurement noise.  The complete Jacobian matrices are provided in Appendix B. 

\begin{figure}[h]
	\centering
	\includegraphics[width=1.0\linewidth]{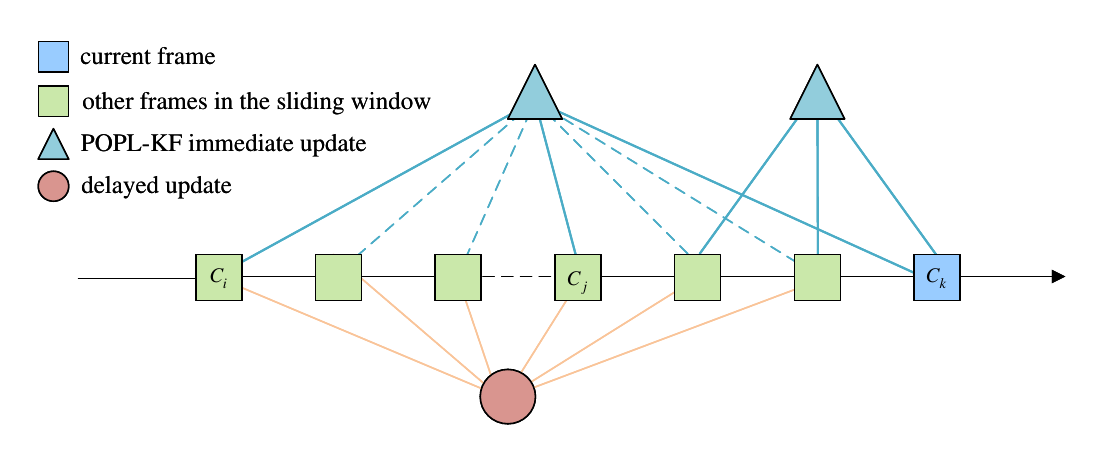}
	\caption{Illustration of immediate and delayed update.}
	\label{fig_3ad}
\end{figure}

By now, we construct the complete pose-only measurement equations for both point and line features. By stacking the measurement equations of all features, we obtain the total measurement matrix, innovation, and noise. Then we can directly calculate the updated system state without null-space projection. The advantages of POPL-KF lie in eliminating the linearization errors of feature 3D positions and enabling the immediate update of visual measurements. The immediate update strategy and the delayed update in MSCKF are shown in Fig.~\ref{fig_3ad}.
\begin{algorithm}[H]
	\caption{Unified Base Frames Selection Algorithm.}
	\begin{algorithmic}[1]
		\REQUIRE Co-observed frame set of point (line) features with $n$ frames.
		\STATE Select the 1-st frame as the base frame $i$, and the $n$-th frame as base-frame $k$;
		\FOR{$j = \{2, \dots, n-1\}$}
		\STATE  Compute the depth (distance) set $\text{D}$  of the point (line) features in the $C_i$ frame, constrained by the image pair $(1,j)$;
		\STATE Calculate the standard deviation $\sigma $, the mean $\mu $ of $\text{D}$, and the coefficient of variation $c=\frac{\sigma }{\mu }$ of $\text{D}$;
		\ENDFOR
		\IF{$c>{{c}_{th}}$}
		\STATE Discard the corresponding unstable features;
		\ELSE
		\FOR{$j = \{2, \dots, n-1\}$}
		\STATE Calculate parallax  ${\varpi }$ and (parallax angle ${\varpi }'$) set using \eqref{deqn_ex47},
		\eqref{deqn_ex48}.
		\ENDFOR
		\STATE  Select the frame corresponding to maximum value in ${\varpi }$ (${\varpi }'$) as the base frame  $j$;
		\ENDIF
		\RETURN Base frame set $(i, j, k)$
	\end{algorithmic}
\end{algorithm}
\subsection{Base-frame Selection Algorithm}
In the pose-only measurement model, the construction of the measurement equations only requires three base-frames. For $n$ images in the sliding window, there exist $C_{n}^{2}$ constraints on direction vectors and distances, and $C_{n}^{3}$ pose-only constraints. Thus, it is essential to devise an effective base-frame selection strategy.

We propose a unified base-frame selection strategy for both point and line features, using parallax and parallax angle between three frames to determine optimal base-frames. Since the inertial navigation system tends to drift over time, we select the 1st observed frame as the first base-frame to reduce cumulative errors and fully exploit the long co-visibility of features. The current frame is chosen as the ${{C}_{k}}$ base-frame to enable immediate updates. The base-frames selection problem is formulated as identifying the optimal frame from the 2nd to the ($n-1$)-th historical frames to construct the most effective constraint on the camera pose. The parallax and parallax angle between the ${{C}_{i}}$ and ${{C}_{j}}$ frames are defined as:
\begin{align}
	\label{deqn_ex44}
	\psi_{ij} &= \left\| \left[ {}^{C_j}\mathbf{f} \right]_{\times} \, {}_{C_i}^{C_j}\mathbf{R} \, {}^{C_i}\mathbf{f} \right\|, \\
	\label{deqn_ex45}
	\psi'_{ij} &= \left\| \arcsin \left( \left[ {}_{C_j}^{C_i}\mathbf{R} \, {}^{C_j}\mathbf{n}_e \right]_{\times} \, {}^{C_i}\mathbf{n}_e \right) \right\|,
\end{align}

As indicated by (17) and (25), when the parallax $\psi_{ij}$ and parallax angle $\psi'_{ij}$ approach zero, the depth and the direction vector in the $C_i$ frame cannot be estimated. The combined parallax and parallax angle of the three base frames are defined as the product of the parallax and parallax angles between each pair of the three frames.
\begin{align}
	\label{deqn_ex47}
	{{\varpi }_{ijk}}    &= {{\psi }_{ij}}{{\psi }_{ki}}{{\psi }_{jk}} \\
	\label{deqn_ex48}
	{{\varpi}^{'}_{ijk}} &= {{\psi}^{'}_{ij}}{{\psi }^{'}_{ki}}{{\psi }^{'}_{jk}},
	\end{align}

To ensure reliable estimation of both the depth and direction vector from the three base-frames, the frame with the maximum combined parallax and angular disparity is selected as the ${{C}_{j}}$ frame. The detailed unified base-frame selection algorithm is presented in Algorithm 1.

\begin{figure*}[htb]
	\centering
	\subfloat[EuRoC]{%
		\includegraphics[width=0.30\textwidth]{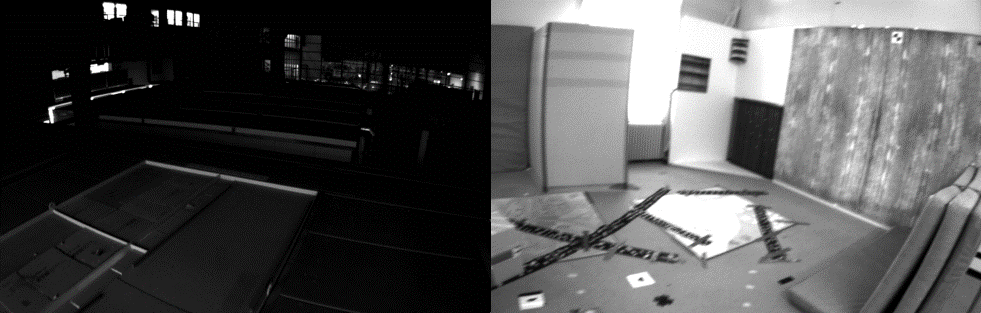}
	}\hspace{0.001\textwidth}
	\subfloat[Kaist]{%
		\includegraphics[width=0.30\textwidth]{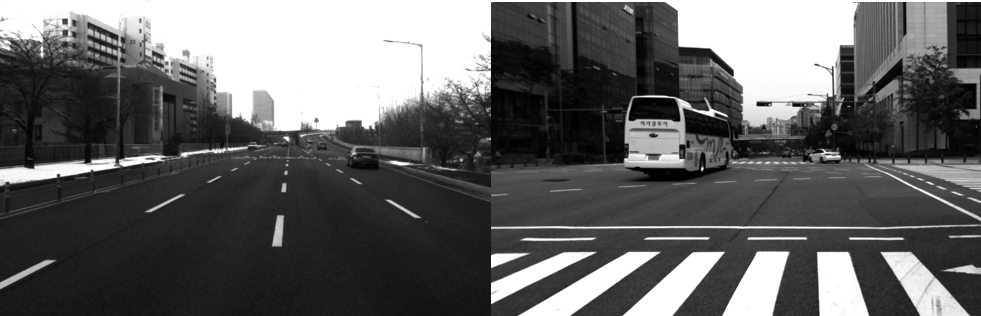}
	}\hspace{0.001\textwidth}
	\subfloat[Corridor]{%
		\includegraphics[width=0.30\textwidth]{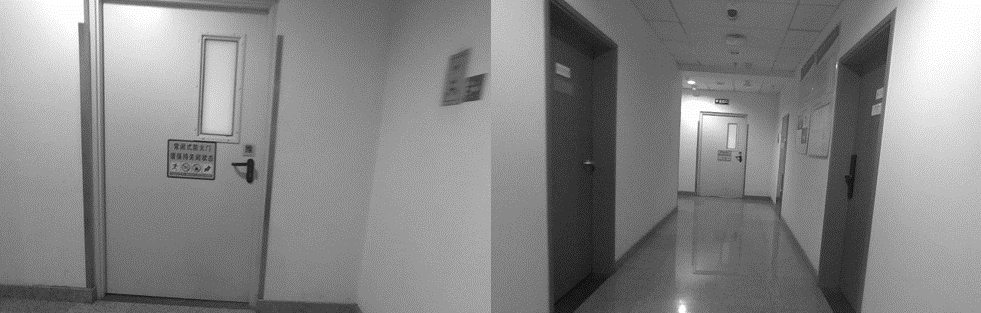}
	}\hspace{0.01\textwidth}
	\\[-0.02cm]
	\subfloat[Lab]{%
		\includegraphics[width=0.30\textwidth]{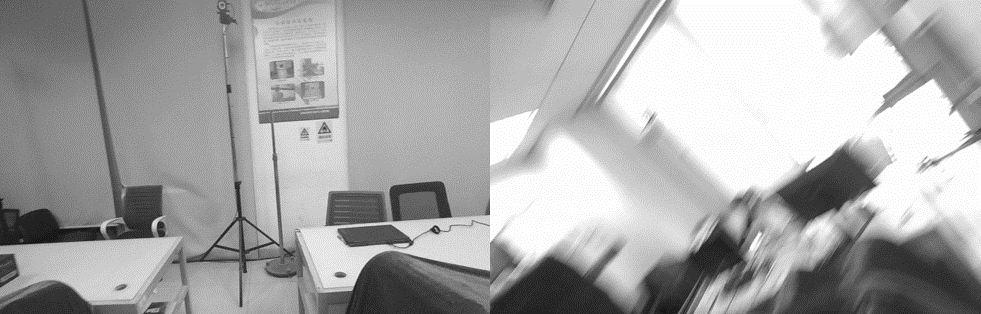}
	}\hspace{0.001\textwidth}
	\subfloat[Campus]{%
		\includegraphics[width=0.30\textwidth]{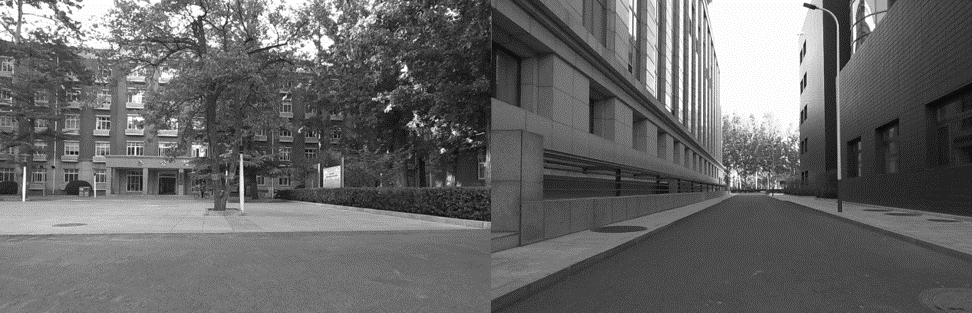}
	}
	\caption{The images sampled from the three datasets.}
	\label{fig_3}
\end{figure*}	
\section{EXPERIMENTS}
\label{exp}
\subsection{Experiments Description}
We extensively evaluate the proposed system on two public datasets and real-world experiments. As shown in Fig.~\ref{fig_3}, the datasets span from indoor environments to complex urban-scale scenes, and include various challenging factors such as low-texture, illumination changes, motion blur, and dynamic objects. To ensure fair comparisons in terms of both accuracy and efficiency, all experiments are conducted on a device equipped with an AMD Ryzen 7-8845H (3.80\, GHz) CPU and 24\,GB of RAM.

\textit{1) Public Dataset:} The EuRoC dataset\cite{burri2016euroc} is collected by a micro-aerial vehicle (MAV) equipped with visual and inertial sensors, comprising two scenarios, an ordinary room and a machine hall. The hall sequences are particularly challenging due to significant illumination variations and weak texture. This dataset provides image data at 20 Hz, IMU measurements at 200 Hz, and ground truth values with millimeter-level accuracy.

The KAIST dataset\cite{jeong2019complex} is collected from vehicles operating in complex urban environments with heavy traffic. Our experiments utilize only the left camera data at 10Hz, industrial-grade MEMS IMU measurements at 100 Hz. We select three sequences, urban32, urban38, and urban39 for evaluation, with trajectory lengths are 7.1km, 11.06 km and 11.42 km, respectively. Since PL-VINS and EPLF-VINS perform poorly on this dataset, they are excluded from the comparison. 

\begin{figure}[H]
	\centering
	\includegraphics[width=0.80\linewidth]{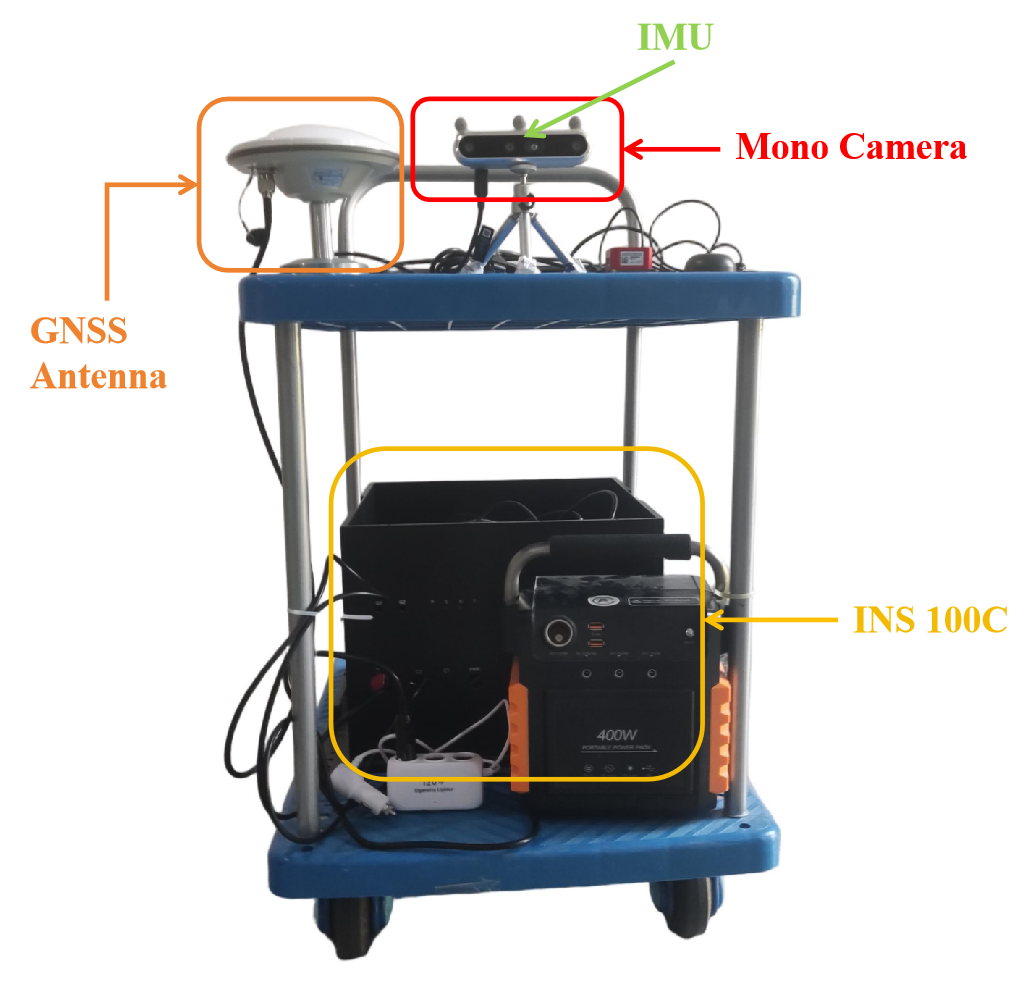}
	\caption{A mobile vehicle.}
	\label{fig_4}
\end{figure}

\textit{2) Real-world experiments:} To further evaluate the proposed system in real-world scenarios, we collect data in both indoor and campus environments at Beihang University using a handheld device and a mobile vehicle. The data collection platform is shown in Fig.~\ref{fig_4}, which includes an Intel RealSense D455 camera (resolution $640\times480$) with a built-in IMU at 200\,Hz, a high-precision IMU (ISA-100C), and a GNSS receiver together with its antenna. All sensors are synchronized, and their extrinsic parameters relative to the IMU are meticulously calibrated. A Trimble Alloy receiver serves as the base station, with its antenna positioned under open-sky conditions. For outdoor experiments, the ground truth trajectory is derived from the NovAtel receiver's tightly integrated navigation output, which provides centimeter-level positioning accuracy.

Based on location and scene complexity, we label the sequences as 1\_corridor, 2\_lab\_easy, 3\_lab\_hard, 4\_campus, and 5\_campus. The ``1\_corridor'' sequence features weak textures and long straight lines. Since ground truth is unavailable, we ensure that the start and end points coincide. The ``3\_lab\_hard'' sequence is characterized by illumination variations and motion blur, while the ``4\_campus'' and ``5\_campus'' include pedestrians and lighting changes. For indoor experiments, ground-truth trajectories are obtained using a motion capture system. For outdoor scenarios, reference trajectories are estimated from smoothed and fused multi-GNSS RTK/IMU (ISA-100C) data using the commercial software Inertial Explorer~(IE) 9.0.

\textit{3) Evaluation Method:} We compare the proposed POPL-KF with PL-VINS, EPLF-VINS, OpenVINS, and PO-KF. Since PO-KF is not open-source, we reproduce it on the SOTA open-source platform OpenVINS. Given that OpenVINS contains the ``SLAM'' feature, the system containing this feature is denoted as ``PO-KF w/SP''. To ensure a fair comparison, we disable the closed-loop module in PL-VINS, EPLF-VINS and evaluate only its odometry performance. All methods are quantitatively evaluated using the EVO tool, with evaluation metrics including absolute translation error (ATE) and absolute rotation error (ARE). In the following subsections, we assess POPL-KF from several aspects: localization accuracy on different datasets, the effectiveness of the line feature selection strategy, robustness under challenging scenarios, and real-time performance.
\subsection{Experiments on Localization Accuracy}

\begin{table*}
	\footnotesize
	\caption{Performance Comparison on the EuRoC and MH Datasets (ATE RMSE in Meters)}
	\label{tab1}
	\centering
	\begin{tabular}{
			>{\centering\arraybackslash}m{2.3cm}
			>{\centering\arraybackslash}m{0.95cm}
			>{\centering\arraybackslash}m{0.95cm}
			>{\centering\arraybackslash}m{0.95cm}
			>{\centering\arraybackslash}m{0.95cm}
			>{\centering\arraybackslash}m{0.95cm}
			>{\centering\arraybackslash}m{0.95cm}
			>{\centering\arraybackslash}m{0.95cm}
		}
		\toprule
		Algorithm & V1\_01 & V1\_02 & V1\_03 & V2\_01 & V2\_02 & V2\_03 & V-Mean \\
		\midrule
		PL-VINS      & 0.061 & 0.086 & 0.144 & 0.066 & 0.099 & \textbf{0.165}\rlap{\textsuperscript{2}} & 0.103 \\
		EPLF-VINS    & 0.076 & 0.078 & 0.138 & 0.162 & 0.078 & 0.221 & 0.126 \\
		OpenVINS     & \textbf{0.057}\rlap{\textsuperscript{1}} & \textbf{0.064}\rlap{\textsuperscript{1}} & \textbf{0.078}\rlap{\textsuperscript{2}} & 0.090 & \textbf{0.055}\rlap{\textsuperscript{1}} & 0.240 & 0.097 \\
		PO-KF        & 0.074 & 0.084 & \textbf{0.067}\rlap{\textsuperscript{1}} & 0.066 & 0.074 & \textbf{0.149}\rlap{\textsuperscript{1}} & \textbf{0.085}\rlap{\textsuperscript{2}} \\
		PO-KF w/SP   & 0.063 & \textbf{0.072}\rlap{\textsuperscript{2}} & 0.083 & \textbf{0.048}\rlap{\textsuperscript{1}} & \textbf{0.066}\rlap{\textsuperscript{2}} & 0.175 & \textbf{0.084}\rlap{\textsuperscript{1}} \\
		POPL-KF      & \textbf{0.060}\rlap{\textsuperscript{2}} & 0.074 & 0.089 & \textbf{0.050}\rlap{\textsuperscript{2}} & \textbf{0.066}\rlap{\textsuperscript{2}} & 0.169 & \textbf{0.085}\rlap{\textsuperscript{2}} \\
		\midrule
		Algorithm & MH\_01 & MH\_02 & MH\_03 & MH\_04 & MH\_05 & MH-Mean & All-Mean \\
		\midrule
		PL-VINS      & 0.271 & 0.297 & 0.124 & 0.279 & 0.214 & 0.237 & 0.164 \\
		EPLF-VINS    & 0.136 & \textbf{0.089}\rlap{\textsuperscript{1}} & \textbf{0.115}\rlap{\textsuperscript{1}} & 0.271 & \textbf{0.175}\rlap{\textsuperscript{2}} & 0.157 & 0.140 \\
		OpenVINS     & 0.141 & 0.153 & 0.132 & 0.185 & 0.361 & 0.194 & 0.142 \\
		PO-KF        & \textbf{0.083}\rlap{\textsuperscript{1}} & 0.149 & 0.203 & 0.173 & 0.329 & 0.187 & 0.132 \\
		PO-KF w/SP   & \textbf{0.084}\rlap{\textsuperscript{2}} & \textbf{0.133}\rlap{\textsuperscript{2}} & 0.188 & \textbf{0.155}\rlap{\textsuperscript{2}} & 0.183 & \textbf{0.148}\rlap{\textsuperscript{2}} & \textbf{0.114}\rlap{\textsuperscript{2}} \\
		POPL-KF      & 0.107 & 0.139 & \textbf{0.123}\rlap{\textsuperscript{2}} & \textbf{0.121}\rlap{\textsuperscript{1}} & \textbf{0.133}\rlap{\textsuperscript{1}} & \textbf{0.125}\rlap{\textsuperscript{1}} & \textbf{0.103}\rlap{\textsuperscript{1}} \\
		\bottomrule
	\end{tabular}
\end{table*}
\begin{figure*}[htb]
	\centering
	\includegraphics[width=0.65\linewidth]{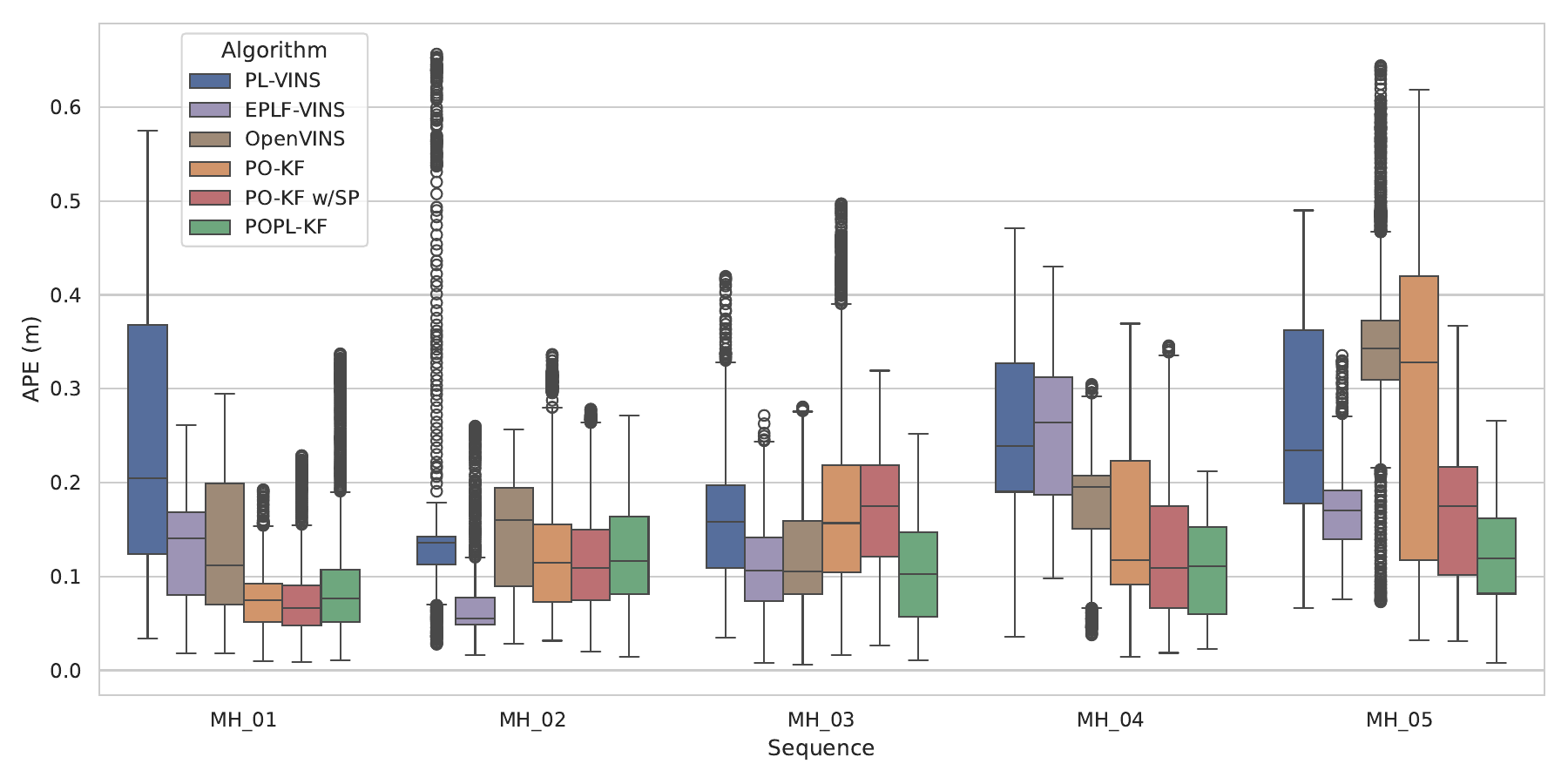}
	\caption{Boxplots of ATE on all MH sequences.}
	\label{fig_5}
\end{figure*}

Table~\ref{tab1} shows the root mean square error (RMSE) of the ATE on the EuRoC dataset. POPL-KF achieves the highest localization accuracy, which is due to the multiple improvements we proposed. In addition, the positioning accuracy of "PO-KF w/SP" is better than that of PO-KF, which shows that SLAM features exhibit low 3D reconstruction error in small-scale scenes, and their incorporation accelerates filter convergence, substantially reducing drift. Furthermore, the comparison between "PO-KF w/SP" and OpenVINS demonstrates that pose-only representation of point features can effectively improve VIO accuracy. 

It should be noted that indoor scenes with limited depth, such as the V sequences, POPL-KF cannot fully exert its advantages, resulting in comparable performance to that of other methods. However, in such scenarios, SLAM features exhibit low 3D reconstruction error, and their incorporation accelerates filter convergence, substantially reducing drift. In contrast, in the larger-scale MH sequences, POPL-KF consistently outperforms OpenVINS in all sequences. Compared to OpenVINS, PO-KF, and ``PO-KF w/SP'', POPL-KF achieves accuracy improvements of 35.57\%, 33.16\%, and 15.54\%, respectively. These results demonstrate that incorporating pose-only representations of line features into VIO can improve positioning accuracy. Meanwhile, the boxplot in Fig.~\ref{fig_5} shows that the proposed pose-only representations for point-line-based VIO achieve the best global consistency in trajectory estimation.

In the MH\_04 and MH\_05 sequences, characterized by low texture and dim lighting, line features exhibit greater robustness than point features, enabling POPL-KF to achieve significantly improved localization accuracy compared to other systems.  These results demonstrate that the proposed method maintains high localization accuracy under visually degraded conditions, confirming its robustness.
\begin{figure}[ht]
	\centering
	\includegraphics[width=0.9\linewidth]{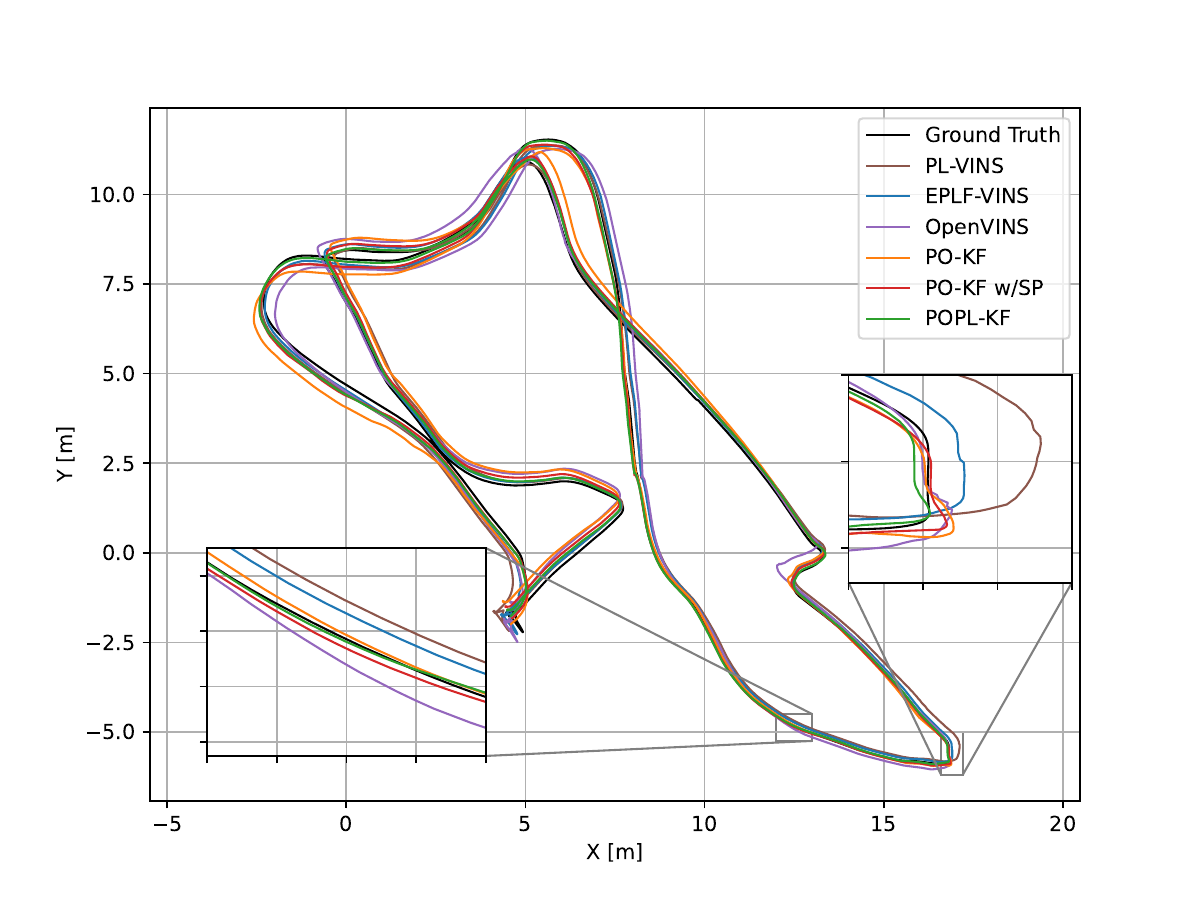}
	\caption{Trajectories produced by PL-VINS, EPLF-VINS, OpenVINS, PO-KF, PO-KF w/SP, and POPL-KF on MH-05 sequence of the EuRoc dataset. The black rectangle highlights the challenging scenarios.}
	\label{fig_6}
\end{figure}
\begin{figure}[ht]
	\centering
	\includegraphics[width=0.9\linewidth]{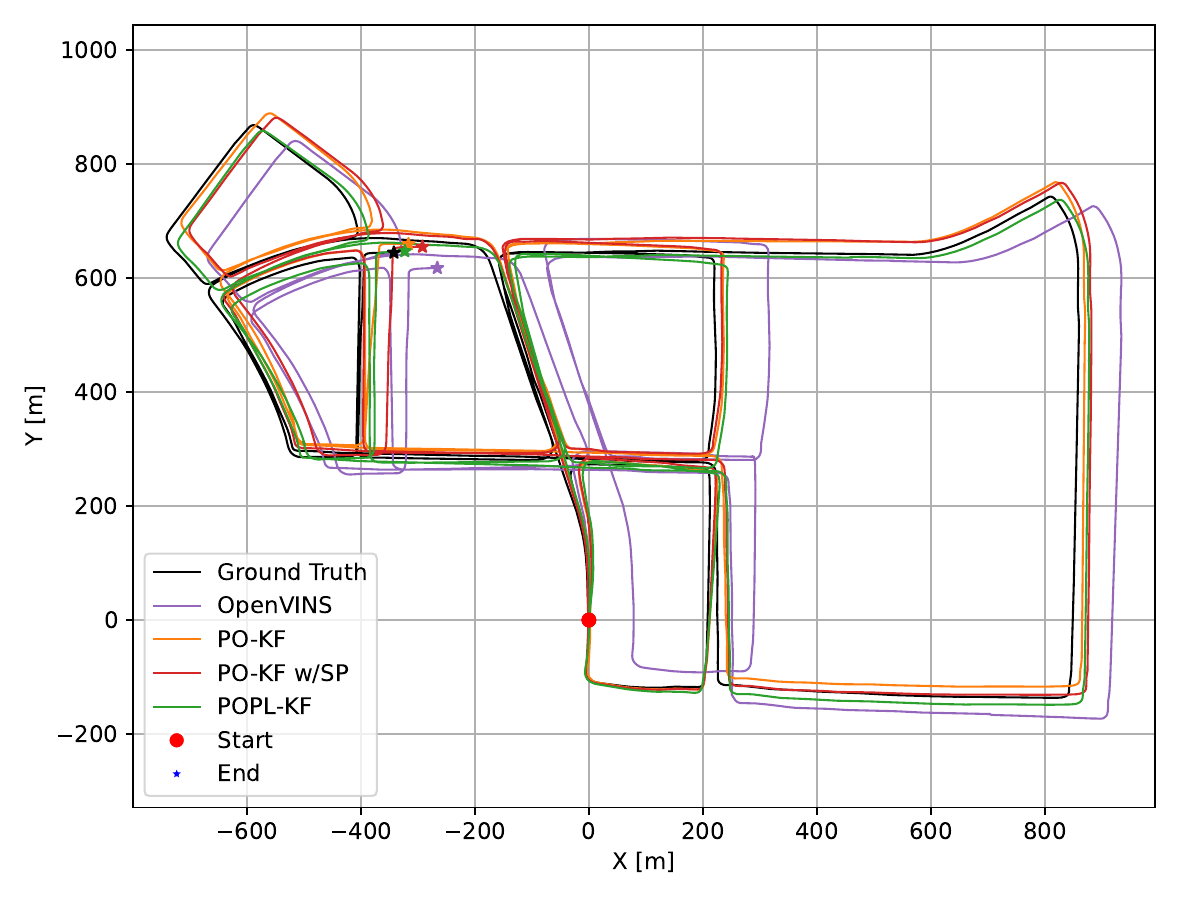}
	\caption{Trajectories produced by OpenVINS, PO-KF, PO-KF w/SP and POPL-KF on urban39 sequence of the KAIST dataset.}
	\label{fig_7}
\end{figure}

Fig.~\ref{fig_6} intuitively shows that POPL-KF produces the trajectory closest to the ground truth on the MH-05 sequence in the XY plane, particularly at sharp turns and under low-light conditions. Moreover, POPL-KF produces smoother trajectories compared to other systems. By explicitly removing feature coordinates from its measurement model, POPL-KF performs immediate updates, effectively suppressing error accumulation and yielding smoother trajectory estimates.

We further evaluate the performance of POPL-KF on the KAIST urban dataset. PL-VINS and EPLF-VINS are excluded from this comparison because their reliance on explicit 3D feature triangulation leads to numerical instability in large-scale environments. Consequently, we compare POPL-KF only with OpenVINS, PO-KF and `` PO-KF w/SP''.  Urban39 is considered one of the most challenging sequences, involving high-speed motion, numerous dynamic objects, and varying exposure conditions. Fig.~\ref{fig_7} presents the trajectories on the urban39 sequence, with all trajectories aligned to the ground truth starting point. OpenVINS exhibits drift on urban39, as such complex urban scenes may degrade visual performance, leading to triangulation failure. In contrast, PO-KF, built on OpenVINS do not require triangulation and enable immediate update, thus maintaining reliable performance under these challenging conditions. The lower positioning accuracy of “PO-KF w/SP'' compared to PO-KF indicates that in large-scale scenes,  the high 3D reconstruction error of SLAM features causes a reduction in accuracy when these features are incorporated. POPL-KF introduces pose-only line features based on the reproduced PO-KF, further improving its accuracy. Under complex urban scenarios, POPL-KF consistently achieves higher trajectory accuracy compared to other systems.

We evaluate the ATE and ARE of OpenVINS, PO-KF, and POPL-KF, with the results summarized in Table~\ref{tab2}. POPL-KF achieves the highest localization accuracy across all test sequences, with improvements of 73.62\% , 28.29\% and 44.33\% compared to OpenVINS, PO-KF and ``PO-KF w/SP'', respectively, demonstrating its practical effectiveness in complex urban environments.  

POPL-KF demonstrates higher positioning accuracy and robustness across both large- and small-scale scenarios, primarily attributed to its pose-only representation and the line feature culling strategy.

\begin{table}[H]
	\caption{ARE/ATE on the Urban Dataset (Unit: deg/s)}
	\label{tab2}
	\centering
	\footnotesize
	\begin{tabular}{
			l
			S[table-format=1.2]@{\,/\,}S[table-format=2.2]
			S[table-format=1.2]@{\,/\,}S[table-format=2.2]
			S[table-format=1.2]@{\,/\,}S[table-format=2.2]
			S[table-format=1.2]@{\,/\,}S[table-format=2.2]
		}
		\toprule
		Sequence & \multicolumn{2}{c}{OpenVINS} & \multicolumn{2}{c}{PO-KF} & \multicolumn{2}{c}{PO-KF w/SP} & \multicolumn{2}{c}{POPL-KF} \\
		\midrule
		\textit{urban32} & 2.02 & 29.03 & 1.68 & 12.29 & 1.80 & 10.89 & \bfseries 1.17 & \bfseries 6.73 \\
		\textit{urban38} & 1.77 & 43.02 & 1.78 & 12.18 & 1.92 & 17.60 & \bfseries 1.49 & \bfseries 11.19 \\
		\textit{urban39} & \bfseries 1.53 & 30.99 & 2.06 & 13.48 & 1.75 & 20.39 & 2.18 & \bfseries 9.29 \\
		\textit{Mean}    & 1.77 & 34.35 & 1.84 & 12.65 & 1.82 & 16.29 & \bfseries 1.61 & \bfseries 9.07 \\
		\bottomrule
	\end{tabular}
\end{table}

\begin{figure}[h]
	\centering
	\subfloat[]{%
		\includegraphics[width=0.85\linewidth]{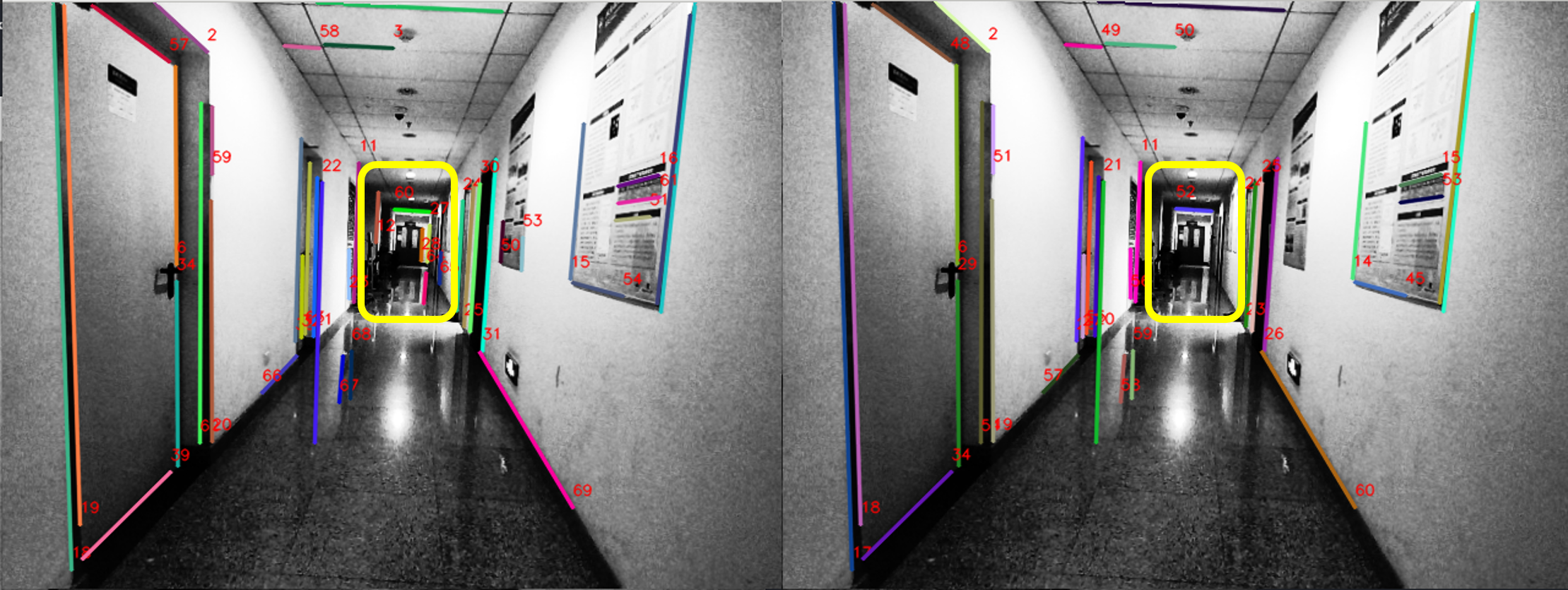}%
		\label{fig:sub1}
	}\par\vspace{-3mm}
	\subfloat[]{%
		\includegraphics[width=0.85\linewidth]{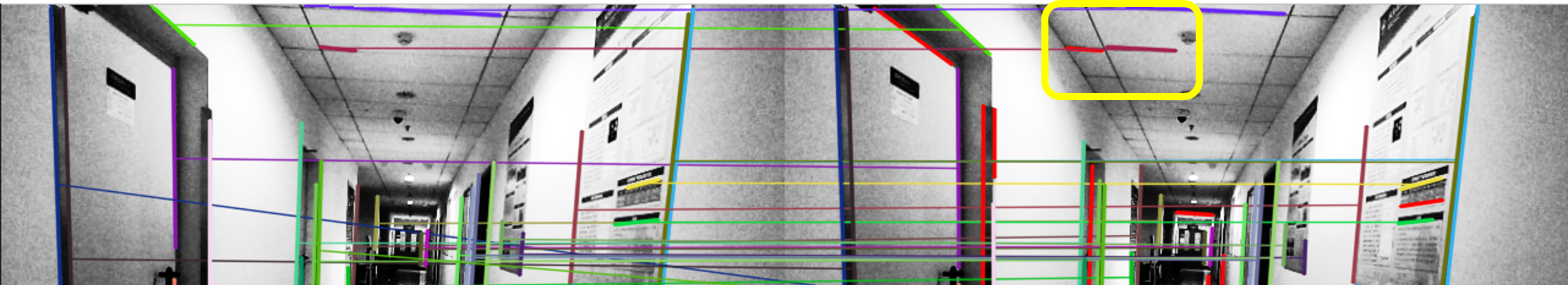}%
		\label{fig:sub2}
	}\par\vspace{-3mm}
	\subfloat[]{%
		\includegraphics[width=0.85\linewidth]{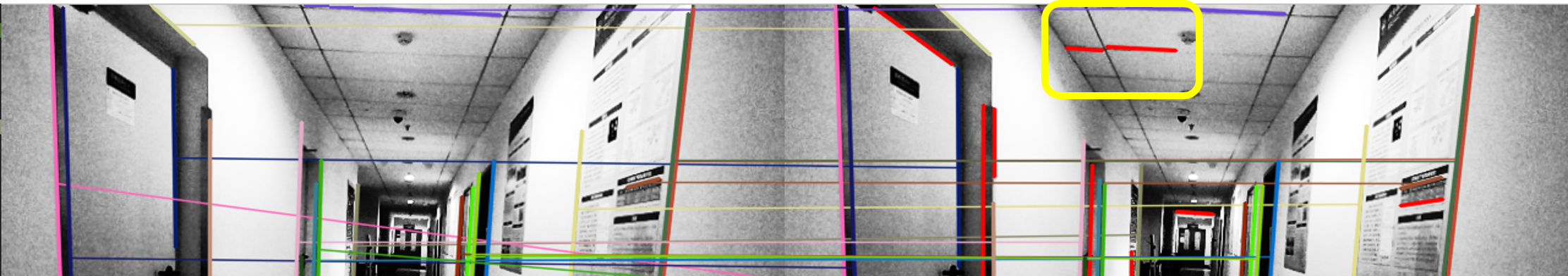}%
		\label{fig:sub3}
	}
	\caption{(a)Demonstration of the grid-based culling strategy, (b) and (c) bidirectional optical flow consistency culling strategy.}
	\label{fig8_ad}
\end{figure}

\subsection{Ablation Experiment}
To validate the effectiveness of the proposed pose-only geometric representation and the line feature culling strategy, we conduct ablation experiments on both the EuRoC and self-collected datasets. As shown in Table~\ref{tab3}, ``POPL-KF w/o cull'' uses the pose-only representation without line feature culling, while ``POPL-KF w/o PO'' updates line features with traditional feature coordinate-based measurements.

As shown on the left side of Fig.~\ref{fig8_ad}(a), the original image contains numerous short and densely clustered line features. These short lines are susceptible to tracking failure and lack persistent temporal association, which in turn degrades the localization accuracy and robustness of the system. After applying the grid-based filtering strategy, as depicted on the right of Fig.\ref{fig8_ad}(a), the distribution of line features becomes more uniform, and redundant segments are effectively suppressed. To further improve observation quality, we introduce a bidirectional optical flow-based filtering strategy. Fig. \ref{fig8_ad}(b) presents the result prior to culling, while Fig. \ref{fig8_ad}(c) shows the result after applying the proposed strategy. The comparison clearly demonstrates that the bidirectional optical flow effectively removes inconsistent line features. The quantitative benefit of the proposed line feature culling strategy is further validated by the results in Table~\ref{tab3}. By removing short, dense, and abnormal line features, the proposed culling strategy improves feature matching quality and enhances system accuracy. 

As shown in Table~\ref{tab3}, POPL-KF outperforms ``POPL-KF w/o PO'', demonstrating the effectiveness of the pose-only representation for line features. This improvement is mainly due to POPL-KF effectively eliminates the linearization errors caused by feature coordinates and immediate measurement updates can be conducted. In the following, we provide a detailed analysis of both factors.

\begin{table*}[htbp]
	\footnotesize
	\caption{ATEs comparison in ablation experiments on the MH sequences of EuRoC dataset (Unit: m)}
	\label{tab3}
	\centering
	\begin{tabular}{
			>{\centering\arraybackslash}m{2.2cm}
			>{\centering\arraybackslash}m{0.8cm}
			>{\centering\arraybackslash}m{0.8cm}
			>{\centering\arraybackslash}m{0.8cm}
			>{\centering\arraybackslash}m{0.8cm}
			>{\centering\arraybackslash}m{0.8cm}
			>{\centering\arraybackslash}m{0.8cm}
		}
		\toprule
		Algorithm & MH\_01 & MH\_02 & MH\_03 & MH\_04 & MH\_05 & Mean \\
		\midrule
		POPL-KF w/o cull & 0.106 & \textbf{0.097} & 0.138 & 0.160 & 0.207 & 0.142 \\
		POPL-KF w/o PO   & \textbf{0.091} & 0.152 & 0.153 & 0.124 & 0.179 & 0.140 \\
		POPL-KF          & 0.107 & 0.139 & \textbf{0.123} & \textbf{0.121} & \textbf{0.133} & \textbf{0.125} \\
		\bottomrule
	\end{tabular}
\end{table*}

\begin{figure}[ht]
	\centering
	\includegraphics[width=0.9\linewidth]{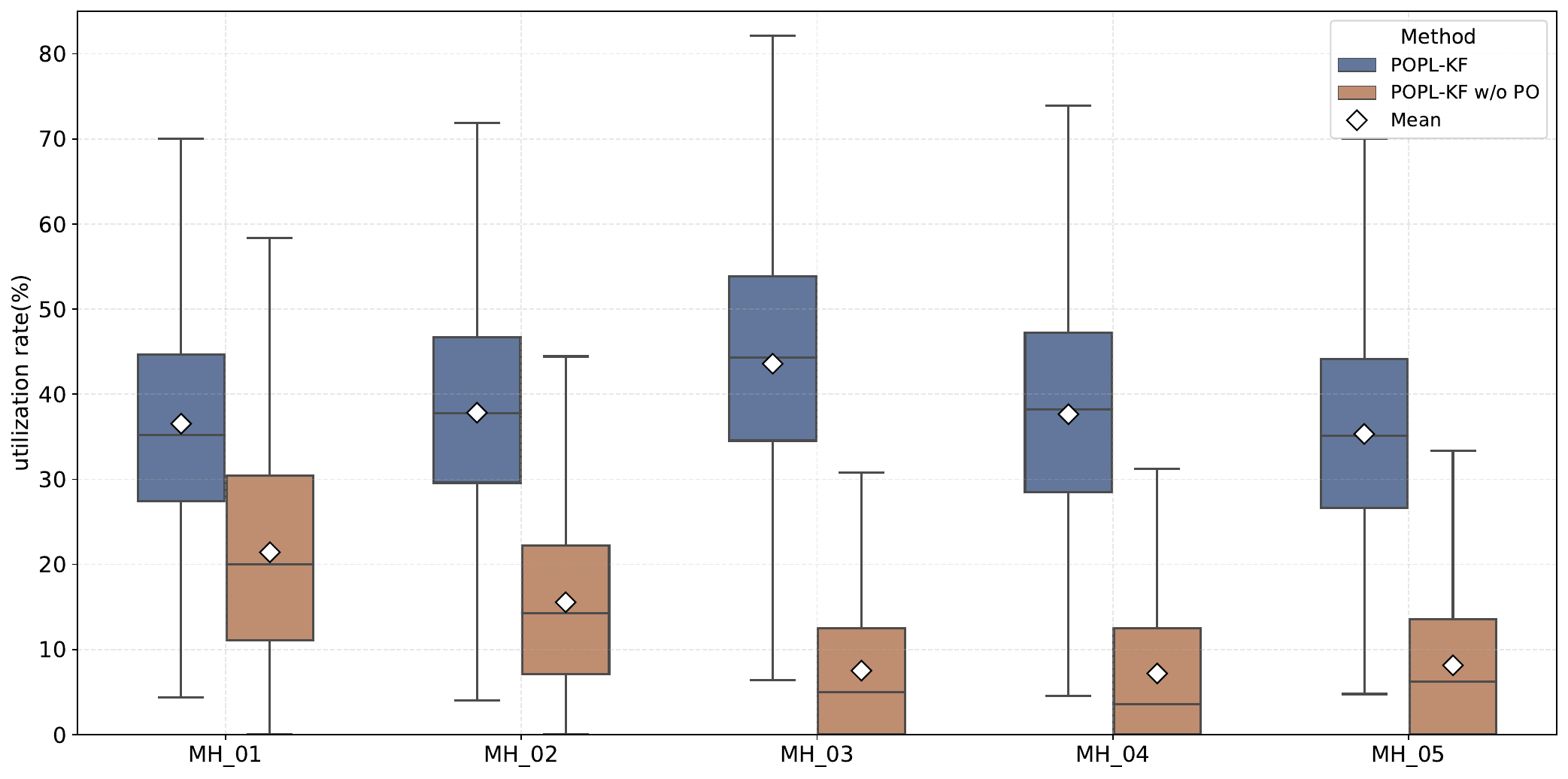}
	\caption{The ratio of the input to the updated measurement counts. }
	\label{fig_8}
\end{figure}
\begin{figure}[ht]
	\centering
	\includegraphics[width=0.9\linewidth]{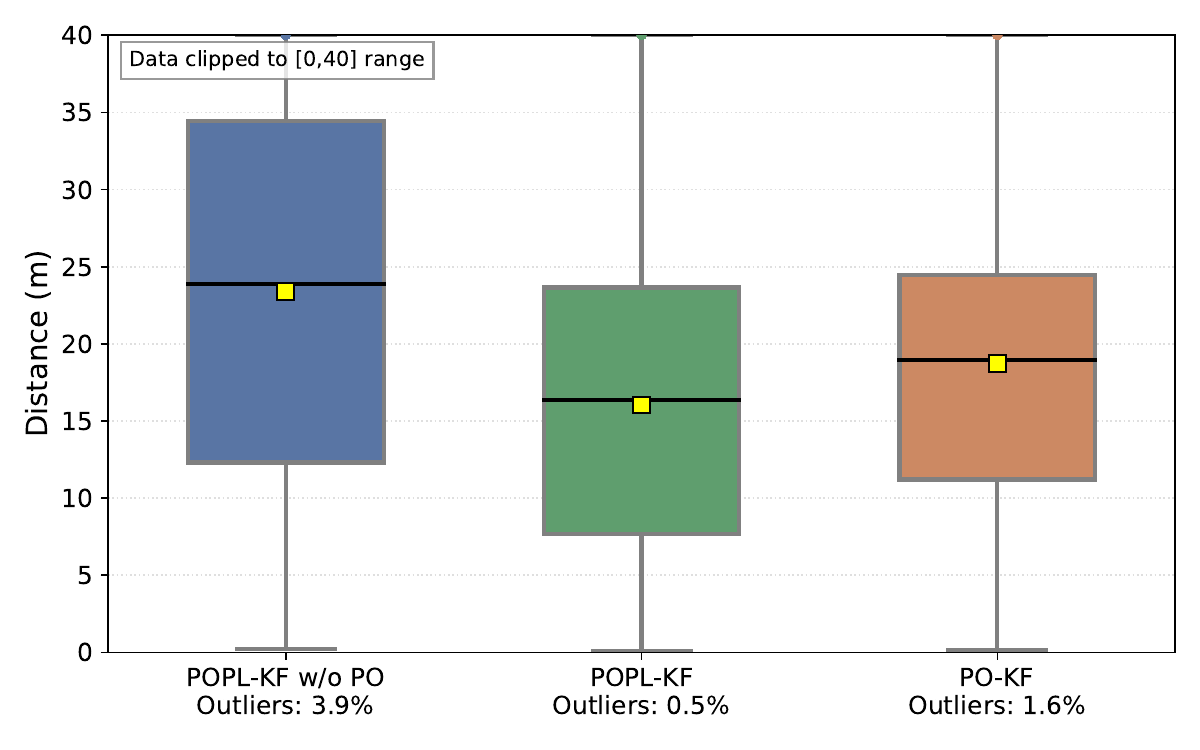}
	\caption{Comparison of estimated distances to line features and point feature depths on the MH-01 sequence of the EuRoC dataset.}
	\label{fig_9}
\end{figure}

Since ``POPL-KF w/o PO'' and POPL-KF share the same visual frontend, the average number of input line feature measurements is similar. Fig.~\ref{fig_8} presents the boxplot of the line feature utilization rate, defined as the ratio of update measurement number to input measurement number.  As shown in Fig.~\ref{fig_8}, POPL-KF exhibits a higher median, overall distribution, and mean utilization rate compared to ``POPL-KF w/o PO'', indicating that it retains more effective visual measurements. ``POPL-KF w/o PO'' performs triangulation only when features are lost or reach their maximum tracking length. Only successfully triangulated features are used to construct the measurement equation, while the rest are discarded, resulting in fewer features being used for updates. In contrast, POPL-KF constructs a measurement equation once a line feature has been tracked for three or more consecutive frames and discards it only when it has been observed for fewer than three frames.

The distance ${{d}_{l}}$, as an intermediate variable in the measurement model, can partially reflect the estimation quality of the line feature coordinates. The depth of point features ${}_{~}^{C}z$ can reflect ${{d}_{l}}$ to a certain extent. Typically, ${{d}_{l}}$ is slightly smaller than ${}_{~}^{C}z$.  Fig.~\ref{fig_9} presents the boxplots of point feature depths in PO-KF and line feature distances in ``POPL-KF w/o PO'' and POPL-KF. As illustrated, the distance estimates of POPL-KF closely align with the depth estimates of PO-KF, with a slightly lower distribution that better matches the expected spatial scale. A significant portion of line features in ``POPL-KF w/o PO'' exhibit large distance estimation errors, with approximately 3.9\% falling outside the expected range. Such errors can cause triangulation failures or substantial estimation inaccuracies, which then propagate through measurement updates, leading to incorrect state corrections and degraded system performance.

In “POPL-KF w/o PO”, the triangulation algorithm lacks the vertical constraint between ${}^{C}\mathbf{v}_e$ and ${}^{C}\mathbf{n}_e$. When the observations are similar, the corresponding coefficient matrix becomes rank-deficient, potentially leading to an ill-conditioned solution.  In contrast, POPL-KF avoids explicit triangulation by using the pose-only representation. The formulation of ${}^{C}\mathbf{v}_{e}$ in \eqref{deqn_ex25} ensures orthogonality with ${}^{C}\mathbf{n}_{e}$, while the non-negativity of ${}^{C}d_{l}$ is guaranteed by \eqref{deqn_ex35}. Moreover, the base-frame selection strategy promotes geometric diversity among frames, which in turn improves numerical stability. The estimation error of ${}^{C}d_{l}$ is further corrected through backpropagated gradient updates, mitigating its impact on the measurement update.

\begin{table*}[ht]
	\footnotesize
	\caption{ATE RMSE on real-word experiments (Unit: m)}
	\label{tab4}
	\centering
	\begin{tabular}{>{\centering\arraybackslash}m{2.5cm} >{\centering\arraybackslash}m{1.2cm} >{\centering\arraybackslash}m{1.2cm} >{\centering\arraybackslash}m{1.2cm} >{\centering\arraybackslash}m{1.2cm} >{\centering\arraybackslash}m{1.2cm} >{\centering\arraybackslash}m{1.2cm}}
		\toprule
		Algorithm & 2\_lab\_easy & 3\_lab\_hard & 4\_campus & 5\_campus & lab\_mean & campus\_mean \\
		\midrule
		EPLF-VINS            & 0.064 & 0.185 & 2.996 & 3.562 & 0.125 & 3.279 \\
		OpenVINS             & 0.070 & 0.154 & 2.786 & 3.044 & 0.112 & 2.915  \\
		PO-KF                & 0.085 & 0.124 & 2.362 & 2.702 & 0.105 & 2.532 \\
		PO-KF w/SP           & \textbf{0.061}\rlap{\textsuperscript{2}} & 0.123 & 2.630 & 3.217 & 0.092 & 2.924 \\
		POPL-KF w/o cull     & 0.067 & \textbf{0.096}\rlap{\textsuperscript{1}} & \textbf{2.115}\rlap{\textsuperscript{2}} & 2.709 & \textbf{0.082}\rlap{\textsuperscript{2}} & \textbf{2.412}\rlap{\textsuperscript{2}}  \\
		POPL-KF w/o PO       & 0.064 & 0.129 & 2.459 & \textbf{2.448}\rlap{\textsuperscript{2}} & 0.097 & 2.454 \\
		POPL-KF              & \textbf{0.044}\rlap{\textsuperscript{1}} & \textbf{0.108}\rlap{\textsuperscript{2}} & \textbf{1.773}\rlap{\textsuperscript{1}} & \textbf{2.266}\rlap{\textsuperscript{1}} &
		\textbf{0.076}\rlap{\textsuperscript{1}} &
		\textbf{2.020}\rlap{\textsuperscript{1}}  \\
		\bottomrule
	\end{tabular}
\end{table*}
\subsection{Real-World Experiments}
\begin{figure*}[htbp]
	\centering
	\begin{minipage}[t]{0.29\textwidth}
		\vspace{0pt}
		\subfloat[1\_corridor]{%
			\includegraphics[width=\linewidth]{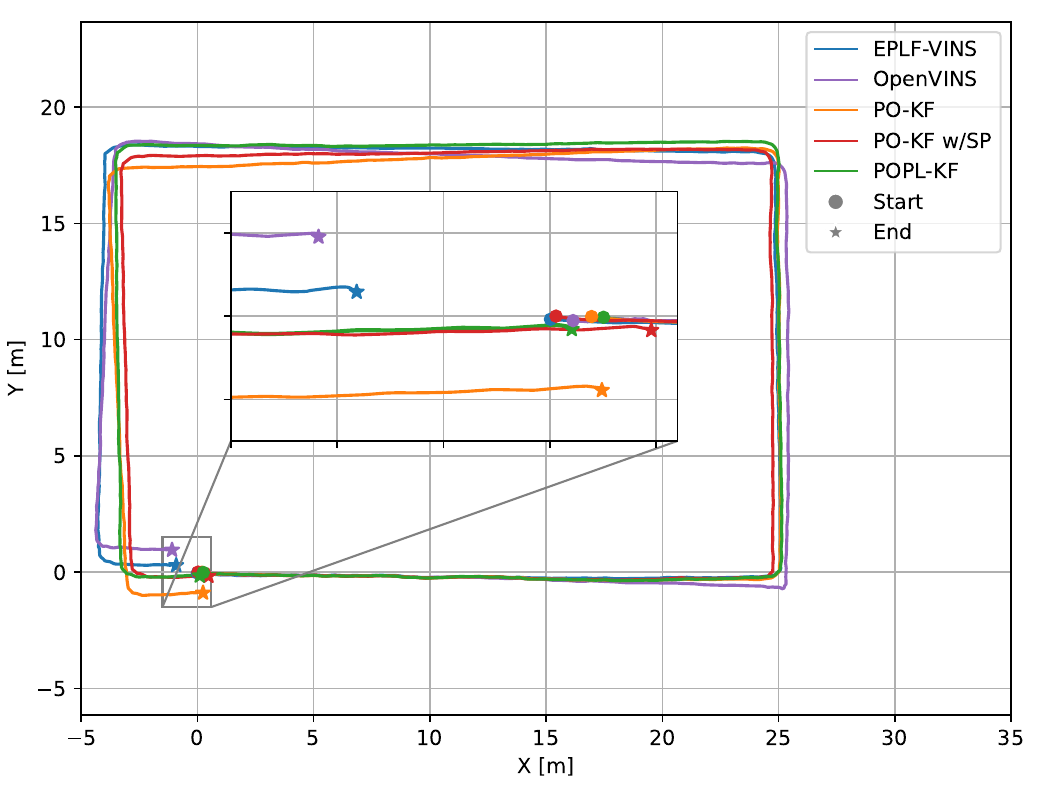}}\\[0.3cm]
		\subfloat[3\_lab\_hard]{%
			\includegraphics[width=\linewidth]{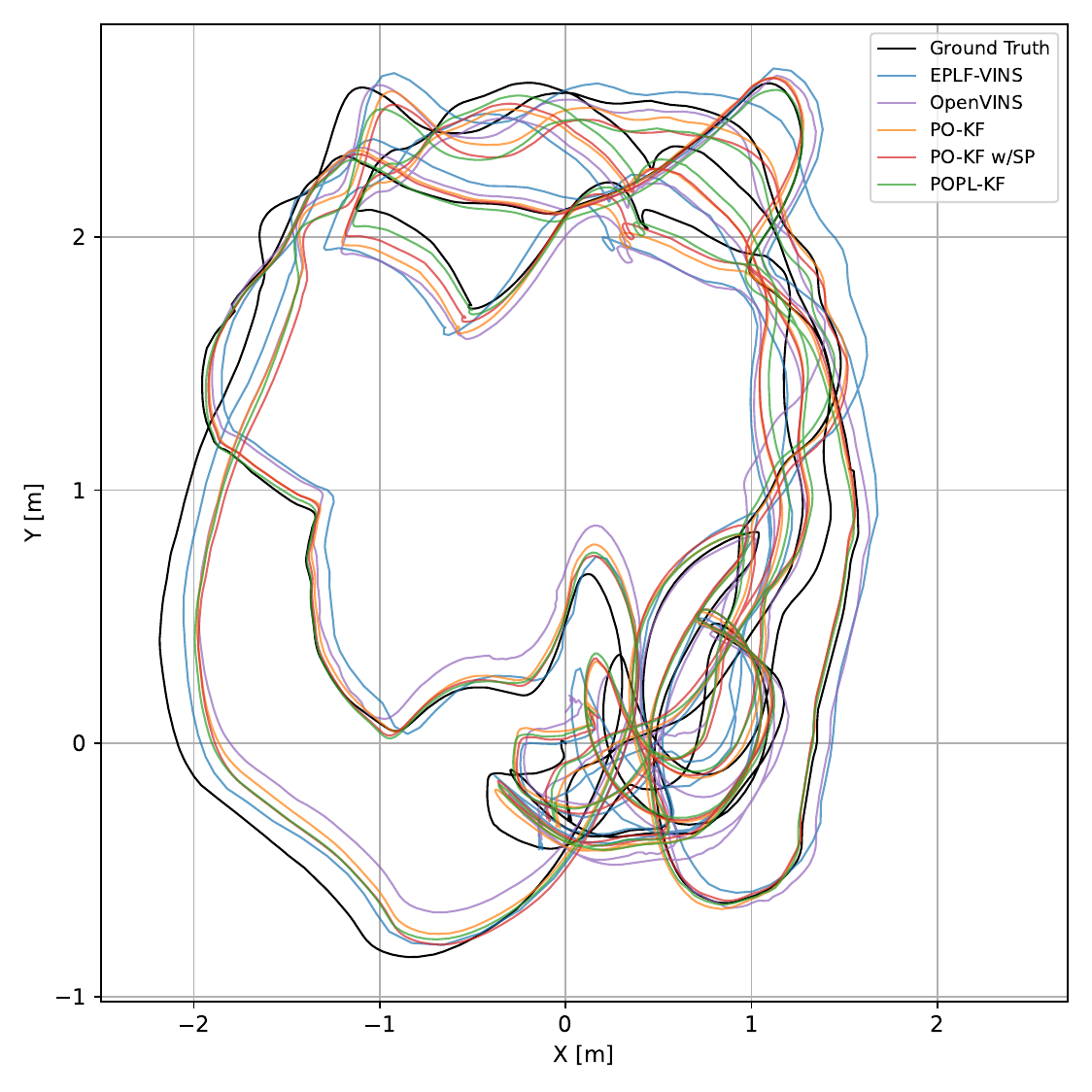}}
	\end{minipage}
	\begin{minipage}[t]{0.29\textwidth}
		\vspace*{-0.1cm} 
		\subfloat[5\_campus]{%
			\includegraphics[width=\linewidth]{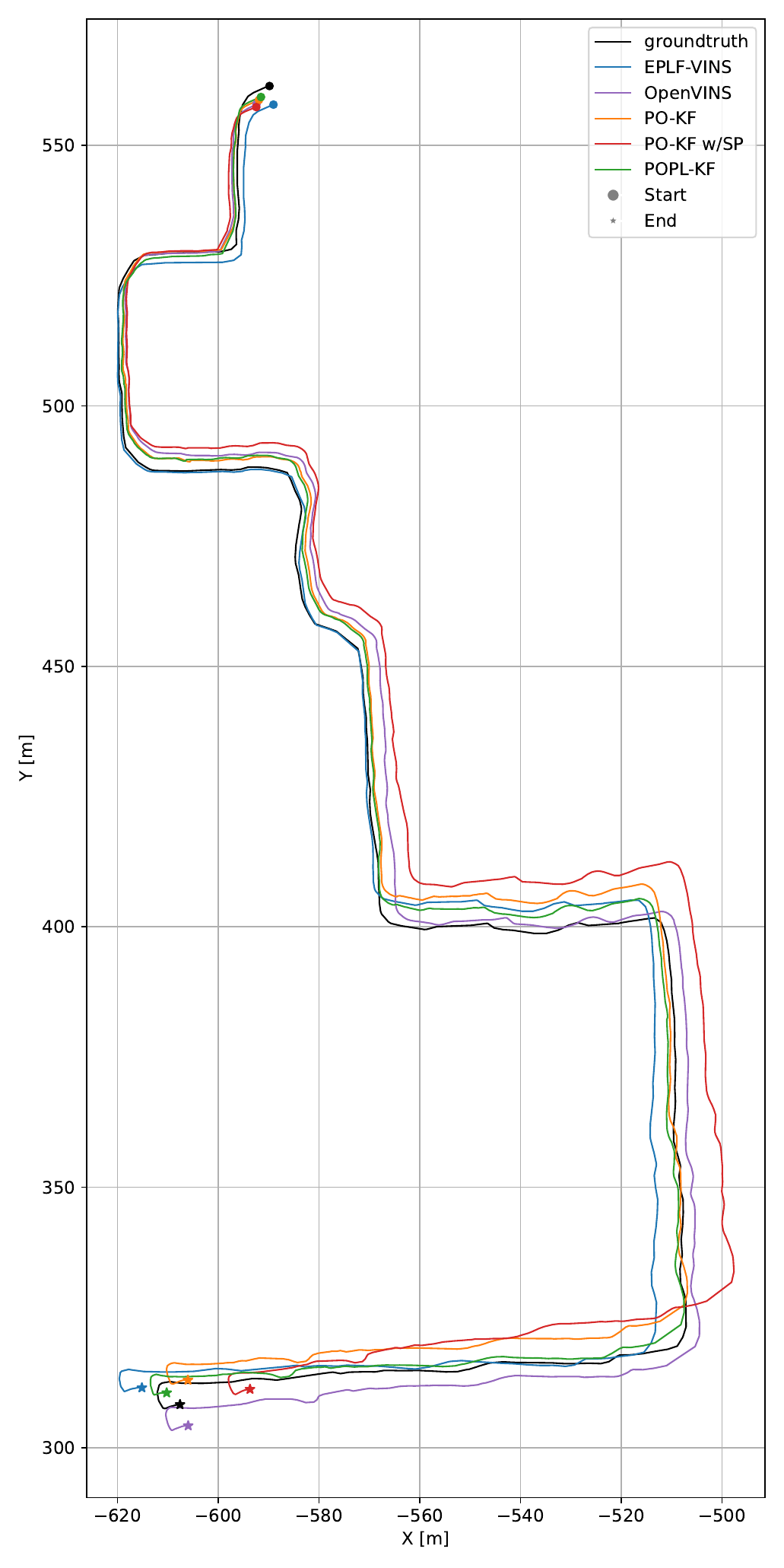}}
	\end{minipage}
	\caption{Estimated trajectories on 1\_corridor, 3\_lab\_hard, and 5\_campus sequences in the XY plane.}
	\label{fig_11}
\end{figure*}

The estimated trajectories for 1\_corridor, 3\_lab\_hard and 4\_campus in the XY plane are shown in Fig.~\ref{fig_11}. Dots indicate starting points, while stars denote endpoints. For clarity, the comparison includes only EPLF-VINS, OpenVINS, PO-KF, ``PO-KF w/SP'', and POPL-KF.

As shown in Fig.~\ref{fig_11}(a), OpenVINS and PO-KF exhibit noticeable drift due to the lack of line feature constraints. As illustrated in Fig.~\ref{fig_3}(c), the scene contains nearly white corridor walls, particularly in the turning areas, making it difficult to extract and match enough stable point features. In contrast, the proposed POPL-KF fully leverages line features and employs a pose-only representation, exhibiting minimal drift in low-texture environments and demonstrating strong robustness in challenging sequences. The ``3\_lab\_hard'' sequence contains motion blur and illumination changes, while ``4\_campus'' includes varying illumination and moving objects. These conditions significantly degrade the performance of feature extraction and matching. Overall, on sequences with available ground truth, POPL-KF produces trajectories that are more consistent with the ground truth, demonstrating higher localization accuracy than the other methods.

Table~\ref{tab4} presents the ATEs of different systems in four real-world sequences. Experimental results demonstrate that incorporating the pose-only representation of line features significantly enhances system accuracy in both indoor and campus environments. The comparisons with ``POPL-KF w/o PO`` and ``POPL-KF w/o cull`` further highlight that the pose-only representation and the line feature culling strategy play a critical role in improving system performance. The average RMSE of POPL-KF in indoor and outdoor scenes is 2.020 m and 0.076 m, respectively, which outperforms all other systems and further illustrates its adaptability and robustness in diverse scenes.

The proposed POPL-KF achieves high precision in both indoor and outdoor real-world environments. Under challenging scenarios such as low texture, motion blur, and illumination changes, POPL-KF consistently outperforms SOTA systems, which validates the effectiveness of the pose-only representation and the line feature culling strategy.

\subsection{Runtime Analysis}
\begin{table*}[ht!] 
	\caption{Detail Runtimes on the MH\_01 sequence of the EuRoC dataset (Unit: ms)}
	\label{tab5}
	\centering
	\begin{tabular}{lcccc}
		\toprule
		Operation & OpenVINS & PO-KF w/SP & POPL w/o PO & POPL-KF \\
		\midrule
		Feature detection \& tracking     & 2.20   & 2.20   & 17.27            & 17.25 \\
		\midrule
		State clone and propagation       & 0.25  & 0.25  & 0.26             & 0.26 \\
		Point update                     & 6.57  & 8.36  & 8.40             & 8.38 \\
		Line update                      & -     & -     & 1.20              & 2.13 \\
		\midrule 
		Marginalization                  & 1.20   & 0.15  & 0.19             & 0.20 \\
		\midrule
		Others                           & 0.15  & 0.14  & 0.17             & 0.18 \\
		\midrule
		Total                           & 10.37 & 11.10 & 27.49            & 28.40 \\
		\bottomrule
	\end{tabular}
\end{table*}

Finally, we evaluate the runtime performance of the proposed POPL-KF system in comparison with the filter-based methods OpenVINS, ``PO-KF w/SP'', and ``POPL-KF w/o PO'' on the MH-01 sequence. As shown in Table~\ref{tab5}, since OpenVINS and ``PO-KF w/SP'', as well as POPL-KF and ``POPL-KF w/o PO'' share the same frontend and state equations, their runtimes for feature tracking and state propagation are nearly identical. As previously mentioned, ``PO-KF w/SP'' and POPL-KF incorporate more update measurements, leading to an increase of 1.79~ms and 0.93~ms in the measurement update runtime. 

Notably, pose-only-based systems avoid feature triangulation, complex Jacobians involving feature coordinates and multiple co-visible poses, and null-space projections. As a result, despite the increased number of measurement updates, the overall runtime of ``PO-KF w/SP'' and POPL-KF increases by only 0.73~ms and 0.91~ms, respectively, which is much smaller than the total per-frame runtime. Although line feature detection and tracking are computationally demanding, POPL-KF maintains real-time performance at over 30~Hz. In summary, it enhances system performance without significantly increasing computational overhead, demonstrating its efficiency in real-time applications.

\section{Conclusion}
\label{con}
In this paper, we propose a pose-only geometric representation for line features and develop a pose-only point and line VIO system POPL-KF. For the first time, both point and line features adopt a pose-only representation. A unified base-frame selection strategy is designed to ensure that the pose-only measurement model has optimal constraints on camera poses. In addition, given the sensitivity of filter-based VIO to outlier observations, we propose a line feature culling method based on image grid partitioning and bidirectional optical flow consistency. The system employs three base-frames to construct measurement equations that explicitly eliminates feature coordinates, which mitigates linearization errors and enables immediate updates. We conduct quantitative and visual evaluations on two public datasets and real-world experiments. Experimental results demonstrate that POPL-KF outperforms the SOTA filter-based methods (OpenVINS, PO-KF) and optimization-based methods (PL-VINS, EPLF-VINS). The real-world experiments cover indoor and outdoor scenarios with challenging conditions such as low texture, illumination changes, motion blur, and dynamic objects. In addition, the proposed system maintains real-time performance at over 30 Hz, indicating its potential for deployment in resource-constrained platforms. In future work, we will focus on replacing the Kalman Filter with the Unscented Kalman Filter to better handle the nonlinear VIO systems and optimizing line feature detection and tracking to further improve computational efficiency.
\appendices
\section{JACOBIAN OF POSE-ONLY MEASUREMENT MODEL FOR POINT FEATURE}
\textit{1) Jacobian of the camera pose to the IMU pose}: As shown in \eqref{deqn_ex23}, the error state of the camera pose is denoted as ${}_{~}^{G}{{\mathbf{\tilde{T}}}_{{{C}_{\eta }}}}={{\left[ \begin{matrix}
			{}_{G}^{{{C}_{\eta }}}{{{\mathbf{\tilde{\theta }}}}^{\text{T}}} & ^{G}\mathbf{\tilde{p}}_{{{C}_{\eta }}}^{\text{T}}  \\
		\end{matrix} \right]}^{\text{T}}},\text{ }\eta =i,\text{ }j,k
$, the Jacobian matrices with respect to the error states of ${}_{~}^{G}{{\mathbf{T}}_{{{I}_{\eta }}}}$ and the extrinsic parameters ${{\mathbf{x}}_{ext}}$ can be defined as
\begin{equation} 
	\label{appendix1} 
	\begin{aligned}
		 \frac{{}_{~}^{G}{{{\mathbf{\tilde{T}}}}_{{{C}_{\eta }}}}}{\partial {}_{~}^{G}{{{\mathbf{\tilde{T}}}}_{{{I}_{\eta }}}}} &=\left[ \begin{matrix}
			{}_{I}^{C}\mathbf{\hat{R}} & {{\mathbf{0}}_{3\times 3}}  \\
			{}_{G}^{{{I}_{\eta }}}{{{\mathbf{\hat{R}}}}^{\text{T}}}{{\left[ {}_{I}^{C}{{{\mathbf{\hat{R}}}}^{\text{T}}}{}_{~}^{C}{{{\mathbf{\hat{P}}}}_{I}} \right]}_{\times }} & {{\mathbf{I}}_{3\times 3}},  \\
		\end{matrix} \right],\\ 
		\frac{\partial {}_{~}^{G}{{{\mathbf{\tilde{T}}}}_{{{C}_{\eta }}}}}{\partial {{{\mathbf{\tilde{x}}}}_{ext}}}&=  \left[ \begin{matrix}
			{{\mathbf{I}}_{3\times 3}} & {{\mathbf{0}}_{3\times 3}}  \\
			{}_{G}^{{{C}_{\eta }}}{{{\mathbf{\hat{R}}}}^{\text{T}}}{{\left[ {}_{~}^{C}{{{\mathbf{\hat{P}}}}_{I}} \right]}_{\times }} & -{}_{G}^{{{C}_{\eta }}}{{{\mathbf{\hat{R}}}}^{\text{T}}}  \\
		\end{matrix} \right],\text{ }\eta =i,j,k.
	\end{aligned}
\end{equation}
\textit{2)  Jacobian of the feature depth to camera poses}: As shown in \eqref{deqn_ex17}, for simplicity,  we define 
 ${\mathbf{A}}_{p} = -{\left[ {}_{~}^{{C_{j}}}\mathbf{f} \right]}_{\times} {}_{G}^{{C_{i}}}\mathbf{R} \left( {}_{~}^{G}{{\mathbf{P}}_{{C_{j}}}} - {}_{~}^{G}{{\mathbf{P}}_{{C_{i}}}} \right)$, 
 ${\mathbf{B}}_{p} = {\left[ {}_{~}^{{C_{j}}}\mathbf{f} \right]}_{\times} {}_{G}^{{C_{j}}}\mathbf{R} {}_{G}^{{C_{i}}}{{\mathbf{R}}^{T}} {}_{~}^{{C_{i}}}\mathbf{f}$. The Jacobian matrices of ${}_{~}^{{C_{i}}}z$  in relation to
 the error states of ${}_{~}^{G}{{\mathbf{T}}_{{C_{i}}}}$ and ${}_{~}^{G}{{\mathbf{T}}_{{C_{j}}}}$ can be formulated as:
\begin{equation}
	\label{appendix_51}
	\begin{aligned}
		\frac{\partial {}_{~}^{{{C}_{i}}}\tilde{z}}{\partial {{{\mathbf{\tilde{A}}}}_{p}}}&= \frac{\mathbf{\hat{A}}_{p}^{\text{T}}}{\left\| {{{\mathbf{\hat{A}}}}_{p}} \right\|\left\| {{{\mathbf{\hat{B}}}}_{p}} \right\|},\frac{\partial {}_{~}^{{{C}_{i}}}\tilde{z}}{\partial {{{\mathbf{\tilde{B}}}}_{p}}} =  \frac{-\mathbf{\hat{B}}_{p}^{\text{T}}}{\left\| {{{\mathbf{\hat{A}}}}_{p}} \right\|{{\left\| {{{\mathbf{\hat{B}}}}_{p}} \right\|}^{3}}}, \\ 
		\frac{\partial {{{\mathbf{\tilde{A}}}}_{p}}}{\partial {}_{~}^{G}{{{\mathbf{\tilde{T}}}}_{{{C}_{i}}}}}&=  \left[ \begin{matrix}
			{{\mathbf{0}}_{3\times 3}} & -{{\left[ {}^{{{C}_{j}}}\mathbf{f} \right]}_{\times }}{}_{G}^{{{C}_{j}}}\mathbf{\hat{R}}  \\
		\end{matrix} \right], \\ 
		\frac{\partial {{{\mathbf{\tilde{A}}}}_{p}}}{\partial {}_{~}^{G}{{{\mathbf{\tilde{T}}}}_{{{C}_{j}}}}}&=  \left[ \begin{matrix}
			-{{\left[ {}_{~}^{{{C}_{j}}}\mathbf{f} \right]}_{\times }}{{\left[ {}_{~}^{{{C}_{j}}}{{{\mathbf{\hat{P}}}}_{{{C}_{i}}}} \right]}_{\times }} & {{\left[ {}_{~}^{{{C}_{j}}}\mathbf{f} \right]}_{\times }}{}_{G}^{{{C}_{j}}}\mathbf{\hat{R}}  \\
		\end{matrix} \right], \\ 
		\frac{\partial {{{\mathbf{\tilde{B}}}}_{p}}}{\partial {}_{~}^{G}{{{\mathbf{\tilde{T}}}}_{{{C}_{i}}}}}&=  \left[ \begin{matrix}
			-{{\left[ {}_{~}^{{{C}_{j}}}\mathbf{f} \right]}_{\times }}{}_{{{C}_{i}}}^{{{C}_{j}}}\mathbf{\hat{R}}{{\left[ {}_{~}^{{{C}_{i}}}\mathbf{f} \right]}_{\times }} & {{\mathbf{0}}_{3\times 3}}  \\
		\end{matrix} \right], \\ 
		\frac{\partial {{{\mathbf{\tilde{B}}}}_{p}}}{\partial {}_{~}^{G}{{{\mathbf{\tilde{T}}}}_{{{C}_{j}}}}}&=  \left[ \begin{matrix}
			{{\left[ {}_{~}^{{{C}_{j}}}\mathbf{f} \right]}_{\times }}{{\left[ {}_{{{C}_{i}}}^{{{C}_{j}}}\mathbf{\hat{R}}{}_{~}^{{{C}_{i}}}\mathbf{f} \right]}_{\times }} & {{\mathbf{0}}_{3\times 3}}  \\
		\end{matrix} \right].  
	\end{aligned}
\end{equation}

\textit{3)  Jacobian of the transformation function}: As shown in \eqref{deqn_ex19}, we derive the Jacobian matrix of $^{{{C}_{k}}}{{\mathbf{p}}_{f}}$ with respect to the error states of ${}_{~}^{G}{{\mathbf{P}}}$ and ${}_{~}^{G}{{\mathbf{T}}_{{{C}_{k}}}}$  as follows:
\begin{equation}
	\begin{aligned}
		& \frac{\partial {}_{~}^{{{C}_{k}}}{{{\mathbf{\tilde{P}}}}_{f}}}{\partial {}_{~}^{G}{{{\mathbf{\tilde{P}}}}_{f}}} = {}_{G}^{{{C}_{k}}}\mathbf{\hat{R}} \text{ }, \\[6pt]
		& \frac{\partial {}_{~}^{{{C}_{k}}}{{{\mathbf{\tilde{P}}}}_{f}}}{\partial {}_{~}^{G}{{{\mathbf{\tilde{T}}}}_{{{C}_{k}}}}} = 
		\left[ \begin{matrix}
			{{\left[ {}_{~}^{{{C}_{k}}}{{{\mathbf{\hat{P}}}}_{f}} \right]}_{\times }} & -  
			{}_{G}^{{{C}_{k}}}\mathbf{\hat{R}} 
		\end{matrix} \right].  
	\end{aligned}
\end{equation}

Given that ${}_{~}^{G}\mathbf{P} ={{\text{h}}_{{}_{~}^{G}\mathbf{P}}}\left( {}^{G}{{\mathbf{T}}_{{{C}_{i}}}},{}_{~}^{{{C}_{i}}}z \right)$, the Jacobian matrices of ${}_{~}^{G}{{\mathbf{P}}}$ concerning the error states of $ {}_{~}^{G}{{{\mathbf{\tilde{T}}}}_{{{C}_{i}}}}$ and ${}_{~}^{{{C}_{i}}}\tilde{z}$ can be defined as
\begin{equation}
	\begin{aligned}
		& \frac{\partial {}_{~}^{G}{{{\mathbf{\tilde{P}}}}_{f}}}{\partial {}_{~}^{G}{{{\mathbf{\tilde{T}}}}_{{{C}_{i}}}}} = 
		\left[ \begin{matrix}
			-{}_{G}^{{{C}_{i}}}{{{\mathbf{\hat{R}}}}^{\text{T}}}{{\left[ {}_{~}^{{{C}_{i}}}\hat{z}{}_{~}^{{{C}_{i}}}\mathbf{f} \right]}_{\times }} & {{\mathbf{I}}_{3\times 3}}  
		\end{matrix} \right], \\[6pt]
		& \frac{\partial {}_{~}^{G}{{{\mathbf{\tilde{P}}}}_{f}}}{\partial {}_{~}^{{{C}_{i}}}\tilde{z}} = {}_{G}^{{{C}_{i}}}{{{\mathbf{\hat{R}}}}^{\text{T}}}{}_{~}^{{{C}_{i}}}\mathbf{f}.  
	\end{aligned}
\end{equation}
\textit{4) Jacobian of the normalized function}: The normalized process is shown in \eqref{deqn_ex20}, the Jacobian matrix of ${}_{~}^{{{C}_{k}}}\mathbf{f}$ with	 regard to the error state of $^{{{C}_{k}}}{{\mathbf{p}}_{f}}$ is expressed as
\begin{equation} 
	\label{appendix_52} 
	\begin{aligned}
		\frac{\partial {}_{~}^{{{C}_{k}}}\mathbf{\tilde{f}}}{{{\partial }^{{{C}_{k}}}}{{{\mathbf{\tilde{p}}}}_{f}}} & = & \left[ \begin{matrix}
			1/{}_{~}^{{{C}_{k}}}\hat{z} & 0 & -{}_{~}^{{{C}_{k}}}\hat{x}/{}_{~}^{{{C}_{k}}}{{{\hat{z}}}^{2}}  \\
			0 & 1/{}_{~}^{{{C}_{k}}}\hat{z} & -{}_{~}^{{{C}_{k}}}\hat{y}/{}_{~}^{{{C}_{k}}}{{{\hat{z}}}^{2}}  \\
			0 & 0 & 0  \\
		\end{matrix} \right].
	\end{aligned}
\end{equation}
\textit{5) Jacobian of the distortion and projection functions}: As defined in \eqref{deqn_ex21}, the distortion function can be defined as :
\begin{equation}
	\label{appendix_49}
	\begin{aligned}
		 {u}'_k={{u}_{k}}(1+{{k}_{1}}{{r}^{2}}+{{k}_{2}}{{r}^{4}})+2{{p}_{1}}{{u}_{k}}{{\nu }_{k}}+{{p}_{2}}({{r}^{2}}+2u_{k}^{2}), \\ 
		 {\nu }'_k={{\nu }_{k}}(1+{{k}_{1}}{{r}^{2}}+{{k}_{2}}{{r}^{4}})+{{p}_{1}}({{r}^{2}}+2\nu _{k}^{2})+2{{p}_{2}}{{u}_{k}}{{\nu }_{k}},
	\end{aligned}
\end{equation}
where $ {{r}^{2}}=u_{k}^{2}+v_{k}^{2}$. The projection process is defined as
\begin{equation}
	\label{appendix_49}
	{\mathbf{\hat{z}}}_{k} =\left[ \begin{matrix}
	{{f}_{x}} & 0 & {{c}_{x}}  \\
	0 & {{f}_{y}} & {{c}_{y}}  \\
	0 & 0 & 1  \\
	\end{matrix} \right]\left[ \begin{matrix}
		{{u}'_k}  \\
		{{\nu }'_k}  \\
		1  \\
	\end{matrix} \right].
\end{equation}

The derivations of the Jacobian matrix of $\mathbf{e}_{k}^{\text{P}}$ with respect to the error state of ${}_{~}^{{{C}_{k}}}\mathbf{f}$ are defined as
\begin{equation}
	\label{appendix_49}
	\frac{\partial \mathbf{e}_{k}^{\text{P}}}{\partial {}_{~}^{{{C}_{k}}}\mathbf{\tilde{f}}} 
	= \left[ \begin{matrix}
		{{f}_{x}} & 0  \\
		0 & {{f}_{y}}  \\
	\end{matrix} \right]
	\left[ \begin{matrix}
		{{k}_{11}} & {{k}_{12}} & 0  \\
		{{k}_{12}} & {{k}_{22}} & 0  \\
	\end{matrix} \right],
\end{equation}
where
\begin{equation}
	\label{appendix_50}
	\small
	\begin{aligned}
		 {{k}_{11}} & =1+{{k}_{1}}{{r}^{2}}+{{k}_{2}}{{r}^{4}}+2{{p}_{1}}{{v}_{k}}+6{{p}_{2}}{{u}_{k}}+4{{k}_{2}}{{r}^{2}}u_{k}^{2}+2{{k}_{1}}u_{k}^{2} \\ 
		{{k}_{12}} & =2{{p}_{1}}{{u}_{k}}+2{{k}_{1}}{{u}_{k}}{{v}_{k}}+4{{k}_{2}}{{r}^{2}}{{u}_{k}}{{v}_{k}}+4{{p}_{2}}{{v}_{k}} \\ 
		{{k}_{21}} & =2{{p}_{2}}{{v}_{k}}+2{{k}_{1}}{{u}_{k}}{{v}_{k}}+4{{k}_{2}}{{r}^{2}}{{u}_{k}}{{v}_{k}}+4{{p}_{1}}{{u}_{k}} \\ 
		{{k}_{22}} & =1+{{k}_{1}}{{r}^{2}}+{{k}_{2}}{{r}^{4}}+2{{p}_{2}}{{u}_{k}}+6{{p}_{1}}{{v}_{k}}+4{{k}_{2}}{{r}^{2}}v_{k}^{2}+2{{k}_{1}}v_{k}^{2}.
	\end{aligned}
\end{equation}

\textit{6) Complete measurement Jacobian matrix}: The complete Jacobian matrices of the point-only pose measurement model are derived based on the chain rule, are defined as:

\begin{equation}
	\label{eq:H_all}
	\begin{aligned}
		\mathbf{H}_{\mathbf{T}_{I_{\eta}}}
		&= \frac{\partial \mathbf{e}_{k}^{\text{P}}}{\partial {}^{G}\tilde{\mathbf{T}}_{I_{\eta}}}
		= \frac{\partial \mathbf{e}_{k}^{\text{P}}}{\partial {}^{G}\tilde{\mathbf{T}}_{C_{\eta}}}
		\cdot \frac{\partial {}^{G}\tilde{\mathbf{T}}_{C_{\eta}}}{\partial {}^{G}\tilde{\mathbf{T}}_{I_{\eta}}}, 
		\quad \eta = i, j, k, \\[0.3em]
		\frac{\partial \mathbf{e}_{k}^{\text{P}}}{\partial {}^{G}\tilde{\mathbf{T}}_{C_{i}}}
		&= \frac{\partial \mathbf{e}_{k}^{\text{P}}}{\partial {}^{C_{k}}\tilde{\mathbf{f}}}
		\cdot \frac{\partial {}^{C_{k}}\tilde{\mathbf{f}}}{\partial {}^{C_{k}}\tilde{\mathbf{P}}_{f}}
		\cdot \frac{\partial {}^{C_{k}}\tilde{\mathbf{P}}_{f}}{\partial {}^{G}\tilde{\mathbf{P}}_{f}} \cdot \bigg( 
		 \frac{\partial {}^{G}\tilde{\mathbf{P}}_{f}}{\partial {}^{G}\tilde{\mathbf{T}}_{C_{i}}}
		+ \\ &\frac{\partial {}^{G}\tilde{\mathbf{P}}_{f}}{\partial {}^{C_{i}}\tilde{z}} \cdot \Big(
	    \frac{\partial {}^{C_{i}}\tilde{z}}{\partial \tilde{\mathbf{A}}_{p}} 
		\cdot \frac{\partial \tilde{\mathbf{A}}_{p}}{\partial {}^{G}\tilde{\mathbf{T}}_{C_{i}}} 
		+ \frac{\partial {}^{C_{i}}\tilde{z}}{\partial \tilde{\mathbf{B}}_{p}} 
		\cdot \frac{\partial \tilde{\mathbf{B}}_{p}}{\partial {}^{G}\tilde{\mathbf{T}}_{C_{i}}}
		\Big) \bigg), \\[0.3em]
		\frac{\partial \mathbf{e}_{k}^{\text{P}}}{\partial {}^{G}\tilde{\mathbf{T}}_{C_{j}}}
		&= \frac{\partial \mathbf{e}_{k}^{\text{P}}}{\partial {}^{C_{k}}\tilde{\mathbf{f}}}
		\cdot \frac{\partial {}^{C_{k}}\tilde{\mathbf{f}}}{\partial {}^{C_{k}}\tilde{\mathbf{P}}_{f}}
		\cdot \frac{\partial {}^{C_{k}}\tilde{\mathbf{P}}_{f}}{\partial {}^{G}\tilde{\mathbf{P}}_{f}}
		\cdot \frac{\partial {}^{G}\tilde{\mathbf{P}}_{f}}{\partial {}^{C_{i}}\tilde{z}} \cdot \Big( \\
		&\quad \frac{\partial {}^{C_{i}}\tilde{z}}{\partial \tilde{\mathbf{A}}_{p}} 
		\cdot \frac{\partial \tilde{\mathbf{A}}_{p}}{\partial {}^{G}\tilde{\mathbf{T}}_{C_{j}}}
		+ \frac{\partial {}^{C_{i}}\tilde{z}}{\partial \tilde{\mathbf{B}}_{p}} 
		\cdot \frac{\partial \tilde{\mathbf{B}}_{p}}{\partial {}^{G}\tilde{\mathbf{T}}_{C_{j}}}
		\Big), \\[0.3em]
		\frac{\partial \mathbf{e}_{k}^{\text{P}}}{\partial {}^{G}\tilde{\mathbf{T}}_{C_{k}}}
		&\!= \!\frac{\partial \mathbf{e}_{k}^{\text{P}}}{\partial {}^{C_{k}}\tilde{\mathbf{f}}}
		\cdot \frac{\partial {}^{C_{k}}\tilde{\mathbf{f}}}{\partial {}^{C_{k}}\tilde{\mathbf{P}}_{f}}
		\cdot \frac{\partial {}^{C_{k}}\tilde{\mathbf{P}}_{f}}{\partial {}^{G}\tilde{\mathbf{T}}_{C_{k}}},\\[0.3em]
		\mathbf{H}_{cam}^{\text{P}} 
		&\!= \!
		\left[
		\frac{\partial \mathbf{e}_{k}^{\text{P}}}{\partial \tilde{k}_1} \!\quad
		\frac{\partial \mathbf{e}_{k}^{\text{P}}}{\partial \tilde{k}_2} \!\quad
		\frac{\partial \mathbf{e}_{k}^{\text{P}}}{\partial \tilde{p}_1} \!\quad
		\frac{\partial \mathbf{e}_{k}^{\text{P}}}{\partial \tilde{p}_2} \!\quad
		\frac{\partial \mathbf{e}_{k}^{\text{P}}}{\partial \tilde{f}_x} \!\quad
		\frac{\partial \mathbf{e}_{k}^{\text{P}}}{\partial \tilde{f}_y} \!\quad
		\frac{\partial \mathbf{e}_{k}^{\text{P}}}{\partial \tilde{c}_x} \!\quad
		\frac{\partial \mathbf{e}_{k}^{\text{P}}}{\partial \tilde{c}_y}
		\right], \\[0.3em]
		&\!= \!
		\left[
		\left[
		\begin{matrix}
			\hat{f}_x u_k r^2 \\
			\hat{f}_y \nu_k r^2
		\end{matrix}
		\right]
		\quad
		\left[
		\begin{matrix}
			\hat{f}_x u_k r^4 \\
			\hat{f}_y \nu_k r^4
		\end{matrix}
		\right]
		\quad
		\left[
		\begin{matrix}
			2\hat{f}_x u_k \nu_k \\
			2\hat{f}_y (r^2 + 2(\nu_k)^2)
		\end{matrix}
		\right]
		\right. \\[0.3em]
		&\quad
		\left.
		\left[
		\begin{matrix}
			\hat{f}_x (r^2 + 2(u_k)^2) \\
			2\hat{f}_y u_k \nu_k
		\end{matrix}
		\right]
		\quad
		\left[
		\begin{matrix}
			u'_k \\
			0
		\end{matrix}
		\right]
		\quad
		\left[
		\begin{matrix}
			0 \\
			\nu'_k
		\end{matrix}
		\right]
		\quad
		\mathbf{0}_{2 \times 2}
		\right], \\[0.8em]
		\mathbf{H}_{ext}^{\text{P}} 
		&= \frac{\partial \mathbf{e}_{k}^{\text{P}}}{\partial \mathbf{\tilde{x}}_{ext}} 
		= \sum_{\eta = i,j,k}
		\frac{\partial \mathbf{e}_{k}^{\text{P}}}{\partial {}^{G}\tilde{\mathbf{T}}_{C_\eta}} 
		\cdot \frac{\partial {}^{G}\tilde{\mathbf{T}}_{C_\eta}}{\partial \mathbf{\tilde{x}}_{ext}},\eta =i,j,k. 
	\end{aligned}
\end{equation}

\section{JACOBIAN OF POSE-ONLY MEASUREMENT MODEL FOR LINE FEATURE}
\textit{1) Jacobian of the camera pose to the IMU pose}: As shown in \eqref{appendix1}.

\textit{2)  Jacobian of the direction vector to camera pose}, As shown in \eqref{deqn_ex32}, that is, the Jacobian matrices of ${}^{{{C}_{i}}}{{\mathbf{v}}_{\alpha }}$ with respect to the error states of ${}_{~}^{G}{{\mathbf{T}}_{{{C}_{i}}}}$ and  ${}_{~}^{G}{{\mathbf{T}}_{{{C}_{k}}}}$ are as shown below:
\begin{equation}
	\begin{aligned}
		\frac{\partial {}^{C_i} \tilde{\mathbf{v}}_{\alpha}}{\partial {}^{G} \tilde{\mathbf{T}}_{C_i}} 
		&= \left[
		\begin{matrix}
			-\operatorname{sgn}(\zeta) [ {}^{C_i} \hat{\mathbf{n}}_e ]_{\times}
			[ {}_{C_j}^{C_i} \hat{\mathbf{R}} {}^{C_j} \hat{\mathbf{n}}_e ]_{\times}
			& \mathbf{0}_{3 \times 3}
		\end{matrix}
		\right], \\[6pt]
		\frac{\partial {}^{C_i} \tilde{\mathbf{v}}_{\alpha}}{\partial {}^{G} \tilde{\mathbf{T}}_{C_j}} 
		&= \left[
		\begin{matrix}
			\operatorname{sgn}(\zeta) [ {}^{C_i} \hat{\mathbf{n}}_e ]_{\times}
			{}_{C_j}^{C_i} \hat{\mathbf{R}}
			[ {}^{C_j} \hat{\mathbf{n}}_e ]_{\times}
			& \mathbf{0}_{3 \times 3}
		\end{matrix}
		\right].
	\end{aligned}
\end{equation}

\textit{3)  Jacobian of the distance vector to camera poses}: As shown in \eqref{deqn_ex34}, for simplicity, let
${{\mathbf{A}}_{l}}= {{\left[ {}_{{{C}_{j}}}^{{{C}_{i}}}\mathbf{R}{}^{{{C}_{j}}}{{\mathbf{n}}_{e}} \right]}_{\times }}{{\left[ {}^{{{C}_{i}}}{{\mathbf{P}}_{{{C}_{j}}}} \right]}_{\times }}{}^{{{C}_{i}}}{{\mathbf{v}}_{\alpha }}$. The Jacobian matrices of ${}^{{{C}_{i}}}{{d}_{l\alpha }}$ in relation to the error states of ${}_{~}^{G}{{\mathbf{T}}_{{{C}_{i}}}}$ and ${}_{~}^{G}{{\mathbf{T}}_{{{C}_{j}}}}$ are expressed as
\vspace{-0.8em}
\begin{equation}
	\begin{aligned}
		\frac{\partial {}^{{C}_{i}}\tilde{d}_{l\alpha}}{\partial \tilde{\mathbf{A}}_{l}} 
		&= \frac{\mathbf{A}_l^{\text{T}}}{\|\mathbf{A}_l\| \cdot \|{}^{{C}_{i}} \mathbf{v}_{\alpha}\|}, \quad
		\frac{\partial {}^{{C}_{i}}\tilde{d}_{l\alpha}}{\partial {}^{{C}_{i}} \tilde{\mathbf{v}}_{\alpha}} 
		= - \frac{\|\mathbf{A}_l\| \cdot {}^{{C}_{i}} \mathbf{v}_{\alpha}^{\text{T}}}{\|{}^{{C}_{i}} \mathbf{v}_{\alpha}\|^3},\\[6pt]
		\frac{\partial \tilde{\mathbf{A}}_l}{\partial {}^{G} \tilde{\mathbf{T}}_{C_i}} 
		&= \left[
		\begin{smallmatrix}
			\Xi \cdot[ {}_{C_j}^{C_i} \hat{\mathbf{R}} {}^{C_j} \hat{\mathbf{n}}_e ]_{\times}
			- [ {}_{C_j}^{C_i} \hat{\mathbf{R}} {}^{C_j} \hat{\mathbf{n}}_e ]_{\times}
			[ {}^{C_i} \hat{\mathbf{v}}_{\alpha} ]_{\times}
			[ {}^{C_i} \hat{\mathbf{P}}_{C_j} ]_{\times} \\
			[ {}_{C_j}^{C_i} \hat{\mathbf{R}} {}^{C_j} \hat{\mathbf{n}}_e ]_{\times}
			[ {}^{C_i} \hat{\mathbf{v}}_{\alpha} ]_{\times}
			{}_{G}^{C_i} \hat{\mathbf{R}}
		\end{smallmatrix}
		\right]^{\text{T}}, \\[6pt]
		\frac{\partial \tilde{\mathbf{A}}_l}{\partial {}^{G} \tilde{\mathbf{T}}_{C_j}} 
		&= \left[
		\begin{matrix}
			\Xi \cdot{}_{C_j}^{C_i} \hat{\mathbf{R}} [ {}^{C_j} \hat{\mathbf{n}}_e ]_{\times}
			- [ {}_{C_j}^{C_i} \hat{\mathbf{R}} {}^{C_j} \hat{\mathbf{n}}_e ]_{\times}
			[ {}^{C_i} \hat{\mathbf{v}}_{\alpha} ]_{\times}
			{}_{G}^{C_i} \hat{\mathbf{R}}
		\end{matrix}
		\right],\\[6pt]
		\frac{\partial \tilde{\mathbf{A}}_l}{\partial {}^{C_i} \tilde{\mathbf{v}}_{\alpha}} 
		&= [ {}_{C_j}^{C_i} \hat{\mathbf{R}} {}^{C_j} \hat{\mathbf{n}}_e ]_{\times}
		[ {}^{C_i} \hat{\mathbf{P}}_{C_j} ]_{\times},
	\end{aligned}
\end{equation}
where, $\Xi =-{{\left[ {{\left[ {}^{{{C}_{i}}}{{{\mathbf{\hat{P}}}}_{{{C}_{j}}}} \right]}_{\times }}{}^{{{C}_{i}}}{{{\mathbf{\hat{v}}}}_{\alpha }} \right]}_{\times }}$.

\textit{4)  Jacobian of the normal vector to camera pose, direction vector and distance vector}: As shown in \eqref{deqn_ex40}, the Jacobian matrices of ${}^{{{C}_{k}}}{{\mathbf{\tilde{n}}}_{l}}$ with respect to the error states of ${}_{~}^{G}{{\mathbf{T}}_{{{C}_{i}}}}$, ${}_{~}^{G}{{\mathbf{T}}_{{{C}_{k}}}}$, ${}^{{{C}_{i}}}{{d}_{l\alpha }}$, and ${}^{{{C}_{i}}}{{v}_{\alpha }}$ are given as follows:
\begin{equation} 
	\label{appendix57}
	\begin{aligned}
		\frac{\partial {}^{C_k} \tilde{\mathbf{n}}_l}{\partial {}_{~}^{G} \tilde{\mathbf{T}}_{C_i}} 
		&= \left[
		\begin{smallmatrix}
			-{}^{C_i} \hat{d}_{l\alpha} \, {}_{C_i}^{C_k} \hat{\mathbf{R}} \left[ {}^{C_i} \hat{\mathbf{n}}_e \right]_{\times}
			- \left[ {}^{C_k} \hat{\mathbf{P}}_{C_i} \right]_{\times} {}_{C_i}^{C_k} \hat{\mathbf{R}} \left[ {}^{C_i} \hat{\mathbf{v}}_{\alpha} \right]_{\times} \\
			-\left[ {}_{C_i}^{C_k} \hat{\mathbf{R}} {}^{C_i} \hat{\mathbf{v}}_{\alpha} \right]_{\times} {}_{G}^{C_k} \hat{\mathbf{R}}
		\end{smallmatrix}
		\right]^T, \\[8pt]
		\frac{\partial {}^{C_k} \tilde{\mathbf{n}}_l}{\partial {}_{~}^{G} \tilde{\mathbf{T}}_{C_k}} 
		&= \left[
		\begin{smallmatrix}
			\left[ {}^{C_i} \hat{d}_{l\alpha} {}_{C_i}^{C_k} \hat{\mathbf{R}} {}^{C_i} \hat{\mathbf{n}}_e \right]_{\times}
			+ \left[ {}^{C_k} \hat{\mathbf{P}}_{C_i} \right]_{\times} \Gamma 
			- \Gamma \left[ {}^{C_k} \hat{\mathbf{P}}_{C_i} \right]_{\times} \\
			\left[ {}_{C_i}^{C_k} \hat{\mathbf{R}} {}^{C_i} \hat{\mathbf{v}}_{\alpha} \right]_{\times} {}_{G}^{C_k} \hat{\mathbf{R}}
		\end{smallmatrix}
		\right]^T, \\[8pt]
		\frac{\partial {}^{C_k} \tilde{\mathbf{n}}_l}{\partial {}^{C_i} \tilde{\mathbf{v}}_{\alpha}} 
		&= \left[ {}^{C_k} \hat{\mathbf{P}}_{C_i} \right]_{\times} {}_{C_i}^{C_k} \hat{\mathbf{R}}, \\[6pt]
		\frac{\partial {}^{C_k} \tilde{\mathbf{n}}_l}{\partial {}^{C_i} \tilde{d}_{l\alpha}} 
		&= {}_{C_i}^{C_k} \hat{\mathbf{R}} {}^{C_i} \hat{\mathbf{n}}_e,
	\end{aligned}
\end{equation}
where, $\Gamma = {{\left[ {}_{{{C}_{i}}}^{{{C}_{k}}}\mathbf{\hat{R}}{}^{{{C}_{i}}}{{{\mathbf{\hat{v}}}}_{\alpha }} \right]}_{\times }}$.

\textit{5) Jacobian of the distortion and projection functions}: As shown in \eqref{deqn_ex41}, the Jacobian matrices of ${}^{{{C}_{k}}}{{\mathbf{\tilde{I}}}_{l}}$ concerning the error states of ${}^{{{C}_{k}}}{{\mathbf{n}}_{l}}$ and $ {{\mathbf{x}}_{\text{cam}}}$ are defined as
\begin{equation} 
	\label{appendix_56} 
	\begin{aligned}
		& \frac{\partial {}^{C_k} \tilde{\mathbf{I}}_l}{\partial {}^{C_k} \tilde{\mathbf{n}}_l}  = \left[
		\begin{matrix}
			\hat{f}_{y} & 0 & 0 \\
			0 & \hat{f}_{x} & 0 \\
			-\hat{f}_{y} \hat{c}_{x} & -\hat{f}_{x} \hat{c}_{y} & \hat{f}_{x} \hat{f}_{y}
		\end{matrix}
		\right],\\[6pt]
		& \frac{\partial {}^{C_k} \tilde{\mathbf{I}}_l}{\partial \tilde{\mathbf{x}}_{\text{cam}}} =
		\left[
		\begin{smallmatrix}
			\mathbf{0}_{1 \times 6} & 0 & n_1 & 0 & 0 \\
			\mathbf{0}_{1 \times 6} & n_2 & 0 & 0 & 0 \\
			\mathbf{0}_{1 \times 6} & -\hat{c}_y n_2 + \hat{f}_y n_3 & -\hat{c}_x n_1 + \hat{f}_x n_3 & -\hat{f}_y n_1 & -\hat{f}_x n_2.
		\end{smallmatrix}
		\right].
	\end{aligned}
\end{equation}

\textit{6) Jacobian of the distance function}: Based on \eqref{deqn_ex42}, we present the Jacobian matrix of $\mathbf{e}_{k}^{\mathbf{L}}$ with respect to the error state of ${}^{{{C}_{k}}}{{\mathbf{I}}_{l}}$ as follows:
\begin{equation}
	\label{appendix_55}
	\frac{\partial \mathbf{e}_{k}^{L}}{\partial {}^{C_k} \tilde{\mathbf{I}}_l} 
	\!=\! \frac{1}{\left( \scriptstyle D \right)^{\scriptstyle {3/2}}} 
	\left[
	\begin{smallmatrix}
		D u_s \!-\! {}^{C_k}\hat{I}_{k1} \, {}^k \bar{\mathbf{s}}^{T} \, {}^{C_k}\hat{\mathbf{I}}_l \\
		D v_s \!-\! {}^{C_k}\hat{I}_{k2} \, {}^k \bar{\mathbf{s}} \, {}^{C_k}\hat{\mathbf{I}}_l \\
		D
	\end{smallmatrix}
	\right.
	\quad
	\left.
	\begin{smallmatrix}
		D u_e \!-\! {}^{C_k}\hat{I}_{k1} \, {}^k \bar{\mathbf{e}}^{T} \, {}^{C_k}\hat{\mathbf{I}}_l \\
		D v_e \!-\! {}^{C_k}\hat{I}_{k2} \, {}^k \bar{\mathbf{e}} \, {}^{C_k}\hat{\mathbf{I}}_l \\
		D
	\end{smallmatrix}
	\right]^T,
\end{equation}
where $D={}^{{{C}_{k}}}\hat{I}_{k1}^{2}+{}^{{{C}_{k}}}\hat{I}_{k2}^{2}$.

\textit{7) Complete measurement Jacobian matrix for line features}: The complete Jacobian matrices of the pose-only measurement model for line features are derived based on the chain rule and are given by
\begin{equation}
    \label{eq:H_all}
	\begin{aligned}
		\mathbf{H}_{\mathbf{T}_{I_\eta}}^{\mathbf{L}} 
		&= \mathbf{H}_{\mathbf{T}_{C_\eta}}^{\mathbf{L}} 
		\cdot \frac{\partial {}^{G} \tilde{\mathbf{T}}_{C_\eta}}{\partial {}^{G} \tilde{\mathbf{T}}_{I_\eta}}, \quad \eta = i,j,k, \\[4pt]
		\mathbf{H}_{\mathbf{T}_{C_i}}^{\mathbf{L}} 
		&= \frac{\partial \mathbf{e}_k^{\mathbf{L}}}{\partial {}^{C_k} \tilde{\mathbf{I}}_l}
		\cdot \frac{\partial {}^{C_k} \tilde{\mathbf{I}}_l}{\partial {}^{C_k} \tilde{\mathbf{n}}_l}
		\cdot \left(
		\frac{\partial {}^{C_k} \tilde{\mathbf{n}}_l}{\partial {}^{G} \tilde{\mathbf{T}}_{C_i}} \right. \\[2pt]
		&\quad 
		+ \frac{\partial {}^{C_k} \tilde{\mathbf{n}}_l}{\partial {}^{C_i} \tilde{d}_{l\alpha}} 
		\left(
		\frac{\partial {}^{C_i} \tilde{d}_{l\alpha}}{\partial \tilde{\mathbf{A}}_l}
		\cdot \frac{\partial \tilde{\mathbf{A}}_l}{\partial {}^{G} \tilde{\mathbf{T}}_{C_i}} \right. \\[2pt]
		&\quad \left. \left.
		+ \left(
		\frac{\partial {}^{C_i} \tilde{d}_{l\alpha}}{\partial \tilde{\mathbf{A}}_l}
		\cdot \frac{\partial \tilde{\mathbf{A}}_l}{\partial {}^{C_i} \tilde{\mathbf{v}}_\alpha}
		+ \frac{\partial {}^{C_i} \tilde{d}_{l\alpha}}{\partial {}^{C_i} \tilde{\mathbf{v}}_\alpha}
		\right)
		\cdot \frac{\partial {}^{C_i} \tilde{\mathbf{v}}_\alpha}{\partial {}^{G} \tilde{\mathbf{T}}_{C_i}}
		\right) \right. \\[2pt]
		&\quad \left.
		+ \frac{\partial {}^{C_k} \tilde{\mathbf{n}}_l}{\partial {}^{C_i} \tilde{\mathbf{v}}_\alpha}
		\cdot \frac{\partial {}^{C_i} \tilde{\mathbf{v}}_\alpha}{\partial {}^{G} \tilde{\mathbf{T}}_{C_i}}
		\right), \\[4pt]
		\mathbf{H}_{\mathbf{T}_{C_j}}^{\mathbf{L}} 
		&= \frac{\partial \mathbf{e}_k^{\mathbf{L}}}{\partial {}^{C_k} \tilde{\mathbf{I}}_l}
		\cdot \frac{\partial {}^{C_k} \tilde{\mathbf{I}}_l}{\partial {}^{C_k} \tilde{\mathbf{n}}_l}
		\cdot \left(
		\frac{\partial {}^{C_k} \tilde{\mathbf{n}}_l}{\partial {}^{C_i} \tilde{d}_{l\alpha}} \right. \\[2pt]
		&\quad \cdot \left(
		\frac{\partial {}^{C_i} \tilde{d}_{l\alpha}}{\partial \tilde{\mathbf{A}}_l}
		\cdot \frac{\partial \tilde{\mathbf{A}}_l}{\partial {}^{G} \tilde{\mathbf{T}}_{C_j}} \right. \\[2pt]
		&\quad \left. 
		+ \left(
		\frac{\partial {}^{C_i} \tilde{d}_{l\alpha}}{\partial \tilde{\mathbf{A}}_l}
		\cdot \frac{\partial \tilde{\mathbf{A}}_l}{\partial {}^{C_i} \tilde{\mathbf{v}}_\alpha}
		+ \frac{\partial {}^{C_i} \tilde{d}_{l\alpha}}{\partial {}^{C_i} \tilde{\mathbf{v}}_\alpha}
		\right)
		\cdot \frac{\partial {}^{C_i} \tilde{\mathbf{v}}_\alpha}{\partial {}^{G} \tilde{\mathbf{T}}_{C_j}}
		\right) \\[2pt]
		&\quad \left.
		+ \frac{\partial {}^{C_k} \tilde{\mathbf{n}}_l}{\partial {}^{C_i} \tilde{\mathbf{v}}_\alpha}
		\cdot \frac{\partial {}^{C_i} \tilde{\mathbf{v}}_\alpha}{\partial {}^{G} \tilde{\mathbf{T}}_{C_j}}
		\right), \\[4pt]
		\mathbf{H}_{\mathbf{T}_{C_k}}^{\mathbf{L}} 
		&= \frac{\partial \mathbf{e}_k^{\mathbf{L}}}{\partial {}^{C_k} \tilde{\mathbf{I}}_l}
		\cdot \frac{\partial {}^{C_k} \tilde{\mathbf{I}}_l}{\partial {}^{C_k} \tilde{\mathbf{n}}_l}
		\cdot \frac{\partial {}^{C_k} \tilde{\mathbf{n}}_l}{\partial {}^{G} \tilde{\mathbf{T}}_{C_k}},\\[4pt]
		\mathbf{H}_{cam}^{\mathbf{L}} &= \frac{\partial \mathbf{e}_{k}^{\mathbf{L}}}{\partial \tilde{\mathbf{x}}_{cam}} 
		= \frac{\partial \mathbf{e}_{k}^{\mathbf{L}}}{\partial {}^{C_{k}} \tilde{\mathbf{I}}_{l}} 
		\cdot \frac{\partial {}^{C_{k}} \tilde{\mathbf{I}}_{l}}{\partial \tilde{\mathbf{x}}_{cam}}, \\[4pt]
		\mathbf{H}_{ext}^{\mathbf{L}} &= \frac{\partial \mathbf{e}_{k}^{\mathbf{L}}}{\partial \tilde{\mathbf{x}}_{ext}} 
		= \sum_{\eta = i,j,k} \mathbf{H}_{\mathbf{T}_{C_{\eta}}}^{\mathbf{L}} 
		\cdot \frac{\partial {}^{G} \tilde{\mathbf{T}}_{C_{\eta}}}{\partial \tilde{\mathbf{x}}_{ext}}.
	\end{aligned}
\end{equation}

\bibliographystyle{IEEEtran}
\bibliography{IEEEabrv,reference}

\end{document}